\documentclass[10pt,journal,compsoc]{IEEEtran}
\usepackage{amsmath,amsfonts}
\usepackage{algorithm}
\usepackage{array}
\usepackage{textcomp}
\usepackage{stfloats}
\usepackage{url}
\usepackage{verbatim}
\usepackage{graphicx}
\usepackage{cite}

\usepackage{algpseudocode}
\usepackage{algorithm}
\usepackage{breqn}

\usepackage{epsfig}
\usepackage{graphicx}
\usepackage{booktabs}
\usepackage{enumitem}
\usepackage{caption}
\usepackage{subcaption}
\usepackage{parskip}
\usepackage{xcolor}

\usepackage{tikz}

\newcommand\submittedtext{%
  \footnotesize This work has been submitted to the IEEE for possible publication. Copyright may be transferred without notice, after which this version may no longer be accessible.}

\newcommand\submittednotice{%
\begin{tikzpicture}[remember picture,overlay]
\node[anchor=south,yshift=10pt] at (current page.south) {\fbox{\parbox{\dimexpr0.65\textwidth-\fboxsep-\fboxrule\relax}{\submittedtext}}};
\end{tikzpicture}%
}

\newcommand{\myparagraph}[1]{\noindent\textbf{#1.}}

\def\onedot{.}

\def\etal{\emph{et al}\onedot}

\begin{document}

\def\method{{UniTransPose}}
\def\methodvit{{EViTPose}}


\title{Transformers with Joint Tokens and Local-Global Attention for Efficient Human Pose Estimation}

\author{Kaleab A. Kinfu~\IEEEmembership{Student Member, IEEE} and René Vidal,~\IEEEmembership{Fellow,~IEEE}
        % <-this % stops a space
\thanks{The authors are with the Center for Innovation in Data Engineering and Science, University of Pennsylvania, Philadelphia, PA, USA. E-mail: \{kinfu, vidalr\}@upenn.edu.}
% <-this % stops a space
% \thanks{}
}

% \markboth{IEEE TRANSACTIONS ON PATTERN ANALYSIS AND MACHINE INTELLIGENCE,~Vol.~##, No.~#, #######~2024}%
% \markboth{IEEE TRANSACTIONS ON PATTERN ANALYSIS AND MACHINE INTELLIGENCE,~Vol.~47, No.~2, Feb.~2025}%


\maketitle

\submittednotice
\begin{abstract}
Convolutional Neural Networks (CNNs) and Vision Transformers (ViTs) have led to significant progress in 2D body pose estimation. However, achieving a good balance between accuracy, efficiency, and robustness remains a challenge. For instance, CNNs are computationally efficient but struggle with long-range dependencies, while ViTs excel in capturing such dependencies but suffer from quadratic computational complexity. This paper proposes two ViT-based models for accurate, efficient, and robust 2D pose estimation. The first one, \textit{\methodvit}, operates in a computationally efficient manner without sacrificing accuracy by utilizing learnable joint tokens to select and process a subset of the most important body patches, enabling us to control the trade-off between accuracy and efficiency by changing the number of patches to be processed. The second one, \textit{{\method}}, while not allowing for the same level of direct control over the trade-off, efficiently handles multiple scales by combining (1) an efficient multi-scale transformer encoder that uses both local and global attention with (2) an efficient sub-pixel CNN decoder for better speed and accuracy.  
% Moreover, by leveraging a unified skeletal representation across multiple datasets,
Moreover, by incorporating all joints from different benchmarks into a unified skeletal representation,
we train robust methods that learn from multiple datasets simultaneously and perform well across a range of scenarios -- including pose variations, lighting conditions, and occlusions. 
Experiments on six benchmarks demonstrate that the proposed methods significantly outperform state-of-the-art methods while improving computational efficiency. \methodvit\ exhibits a significant decrease in computational complexity (30\% to 44\% less in GFLOPs) with a minimal drop of accuracy (0\% to 3.5\% less), and \method\ achieves accuracy improvements ranging from 0.9\% to 43.8\% across these benchmarks.

\end{abstract}


\section{Introduction}

\IEEEPARstart{U}{nderstanding} humans in images and videos has played a central role in computer vision for several decades. Human pose estimation, which involves detecting human figures and determining their poses, has found numerous applications in action recognition~\cite{Du2015HierarchicalRN, Yan2018SpatialTG}, motion analysis~\cite{Stenum2020TwodimensionalVA}, gaming~\cite{Ke2010RealTime3H}, video surveillance~\cite{Lamas2022HumanPE}, human-computer interaction~\cite{Ke2018ComputerVF}, virtual and augmented reality~\cite{Obdrzlek2012RealTimeHP, Lin2010AugmentedRW}, and more recently also in healthcare~\cite{Melcio2021DeepRehabRT, Li2020HumanPE, Kinfu2025ComputerizedAO}.

\myparagraph{Desiderata for Human Pose Estimation Networks} 
An ideal human pose estimation network should:

\begin{itemize}[]

\item Accurately locate body joints, even in complex scenes with multiple interactions.

\item Ensure computational efficiency to
make the solution viable for resource-constrained applications.

\item Demonstrate robustness across a range of scenarios, including different scales, lighting conditions, occlusions, and backgrounds. 

\end{itemize}
 
However, state-of-the-art methods struggle to strike a good balance among accuracy, efficiency, and robustness, primarily due to constraints in their encoding and decoding approaches, as well as the inconsistency in dataset annotations.

\myparagraph{Challenges in Encoding Approaches}
Most recent works in 2D pose estimation employ encoder-decoder architectures based on Convolutional Neural Networks (CNNs). CNN-based encoders perform well on low- to mid-resolution images, but their performance deteriorates in high-resolution images due to the inability of CNNs to capture long-range dependencies among image regions. 
Vision Transformers (ViTs) \cite{ Khan2022TransformersIV} have recently emerged as powerful alternatives to CNNs for solving various computer vision tasks. ViTs use Multi-head Self-Attention (MSA) to capture long-range dependencies among patch tokens and thus produce a global representation of the image. 
Several works~\cite{Yang2021TransPoseKL, Li2021TokenPoseLK, Yuan2021HRFormerHT, Xu2022ViTPoseSV} have utilized ViTs for human pose estimation and demonstrated its effectiveness. Nevertheless, the computational complexity of ViTs, which scales quadratically with the number of input tokens, presents a significant challenge for processing high-resolution images, making ViTs less feasible for practical use in resource-constrained environments.

Another challenge in ViTs is that the fixed scale of patches is not ideal for dense prediction tasks where the visual elements are of variable scale. Multi-scale transformers~\cite{Liu2021SwinTH, Dong2022CSWinTA, Wang2021PyramidVT} address both the fixed scale and computational complexity issues by constructing hierarchical feature maps and restricting the computation of self-attention to a local window, thus achieving a linear complexity with respect to the number of patches. However, the use of local attention limits the ability of the network to capture long-range dependencies among tokens~\cite{Liu2021SwinTH}.

\myparagraph{Challenges in Decoding Approaches} 
Current human pose estimation methods typically employ one of two decoding approaches: direct key-point regression or heat-map decoding. In key-point regression, the model directly predicts the coordinates of each joint in the image, whereas in heat-map decoding the model generates a heat map whose highest values indicate the coordinates of the joints. 
Although key-point regression is more efficient than heat-map decoding, the latter is more accurate and robust to occlusions~\cite{Yu2020heatmapRV,Tompson2014JointTO,Nibali2018NumericalCR}.

\myparagraph{Dataset Annotation Inconsistencies} 
Training a single, unified model across all datasets can enhance both accuracy and robustness as it allows the model to learn from a wide range of human pose variations under numerous conditions, ultimately leading to better generalization. Moreover, it can simplify the training process by eliminating the need to train separate models on each dataset--a common practice among state-of-the-art methods that is time-consuming and resource-intensive. Although there are several large- and small-scale human pose estimation benchmarks~\cite{Lin2014MicrosoftCC, Wu2017AIC, Vendrow2022JRDBPoseAL, Andriluka20142DHP, Li2018CrowdPoseEC, Zhang2019Pose2SegDF} we can utilize to train a unified model, the variability in the number and location of annotated joints~\cite{Zheng2019DeepLH} across these benchmarks complicates the training process.

\myparagraph{Paper Contributions}
In this paper, we address the aforementioned challenges by proposing two vision transformer-based models. The first model, \methodvit, offers a direct control of the trade-off between accuracy and efficiency by carefully selecting a subset of patches to be processed. The second model, \method, improves robustness to scale changes while offering a good balance between accuracy and efficiency, by using an efficient multi-scale transformer architecture.
We also propose a unified skeletal representation that improves model training across multiple datasets, boosting performance and generalization. 
Specifically, this paper makes the following contributions:
\begin{itemize}[]

\item \emph{\methodvit}\ selects and processes a subset of patches containing the most important information about body joints. Specifically, we use learnable joint tokens that explicitly learn the joint embeddings to identify patches that are more likely to include the true joints. This method improves efficiency while maintaining high accuracy, demonstrating a 30\% to 44\% reduction in computational complexity with a minimal accuracy drop of 0\% to 3.5\% across six benchmarks.

\item \emph{\method}\ 
integrates an efficient multi-scale transformer encoder with an efficient sub-pixel CNN-based decoder to achieve better accuracy and efficiency. Notably, the encoder uses a \emph{local-global attention mechanism} that includes (i) local patch-to-patch, (ii) global patch-to-joint, and (iii) local joint-to-patch attention mechanisms, the last two using learnable joint tokens.
\method\ improves the accuracy of state-of-the-art methods by 0.9\%-2.4\% on MS-COCO, 3.3\%-5.7\% on AI Challenger, and 6.2\%-43.8\% on OCHuman benchmarks.

\item By utilizing both key-point regression and heat-map decoding, \methodvit\ and \method\ are designed to be flexible, providing a choice between efficiency and accuracy. The combination of both decoders offers the best of both worlds and is more adaptable to different use cases. For instance, key-point regression could be used for quick estimates, while heat-map decoding could be used for more accurate predictions as needed. 

\item By using a \emph{unified skeletal representation} that incorporates all joints from multiple datasets to train our methods, \methodvit\ and \method\ learn to be robust to different number of joints and annotation styles.

\item Comprehensive experiments on six commonly used 2D human pose estimation benchmarks (i.e. MS-COCO, MPII, AI Challenger, JRDB-Pose, CrowdPose, and OCHuman) demonstrate that our proposed methods trained on a unified skeletal representation across multiple datasets significantly improve the trade-off among accuracy, computational efficiency, and robustness compared to state-of-the-art approaches.
\end{itemize}

\section{Related Work}
In this section, we briefly review single-  and multi-person 2D pose estimation approaches and CNN- and ViT-based methods related to this work.

\subsection{Single-Person Pose Estimation}

As the name suggests, single-person pose estimation techniques assume there is one person in the image. Deep learning techniques for single-person pose estimation generally fall into two primary categories: key-point regression-based and heat-map-based approaches.  
In the key-point regression-based approach, the task is treated as a direct joint location regression problem and the network learns a mapping from the input image to the coordinates of body joints within the image via end-to-end training, as exemplified by the pioneering work in DeepPose~\cite{Toshev2014DeepPoseHP}. This approach is favored for its computational efficiency since it doesn't need the intermediate step of heat map generation and directly targets joint coordinates. As a result, it can be faster in terms of computation time, making it suitable for applications requiring real-time performance. However, despite its efficiency, the key-point regression approach typically suffers from reduced robustness and lower accuracy, particularly in complex scenarios, such as challenging poses, under occlusion, or when the subject's appearance varies significantly. These limitations stem from the direct regression task, which may not capture the subtleties of spatial relationships.

In contrast, the heat map-based approach is designed to predict the approximate coordinates of the joints encoded via 2D Gaussian heat maps that indicate the probability of a joint's presence at each pixel location, with the peak of the heat map centered on the joint location. This method converts the pose estimation problem into a spatial probability distribution task, allowing the model to learn and predict the likelihood of joint positions across the entire image area. This approach is generally more robust than key-point regression. 
This robustness is attributed to the heat map's ability to capture and integrate subtle cues across the image, leading to more accurate joint detection in visually complex situations as shown in several works~\cite{Yu2020heatmapRV, Tompson2014JointTO, Nibali2018NumericalCR}.

Our work incorporates both decoding approaches to leverage their respective strengths. This offers flexibility to balance the efficiency and accuracy trade-off according to the user's specific needs. While our primary focus is on the heat-map-based decoding method due to its superior accuracy, including the key-point regression-based method allows our models to adapt to a wide range of applications.

\subsection{Multi-Person Pose Estimation}
The multi-person pose estimation task is more challenging than single-person pose estimation because it must determine the number of people and associate detected joints with the correct person. Multi-person pose estimation techniques can be divided into top-down and bottom-up approaches. Top-down approaches~\cite{Sun2019DeepHR, Xiao2018SimpleBF, Newell2016StackedHN, Yang2021TransPoseKL, Yuan2021HRFormerHT} use generic person detectors to extract a set of boxes from the input images, each of which belongs to a single person. These methods then apply single-person pose estimators to each person box to produce multi-person poses, often resulting in high accuracy per detected person. However, it can be computationally expensive, especially as the number of people increases. Conversely, the bottom-up approaches~\cite{Cao2021OpenPoseRM, Newell2017AssociativeEE, Cheng2020HigherHRNetSR, Shi2022EndtoEndMP} first identify all the body joints in a single image and then groups them for each person in the scene. This approach tends to be more efficient, particularly in crowded scenes, as it does not require individual person detection before pose estimation. The trade-off, however, is that the accuracy of such approaches can be lower, particularly in complex scenes. In this work, we will follow the top-down approach.

\subsection{Convolutional Neural Network Based Approaches}

Convolutional Neural Networks (CNNs) have been extensively used for human pose estimation and achieved high performance.
% ~\cite{Sun2019DeepHR, Cao2021OpenPoseRM, Xiao2018SimpleBF, Newell2016StackedHN, Chen2018CascadedPN}. 
Notably, the work by Toshev and Szegedy~\cite{Toshev2014DeepPoseHP} introduced DeepPose, a pioneering approach utilizing CNNs to directly regress body joint coordinates, marking a paradigm shift towards deep learning-based methods in pose estimation. 
The performance of such approaches has been improved by employing multi-stage architectures, stacking deeper blocks and maintaining high-resolution and multi-scale representations. Wei~\etal\cite{Wei2016ConvolutionalPM} introduced Convolutional Pose Machines, which iteratively refines predictions through a multi-stage architecture. Concurrently, Newell~\etal\cite{Newell2016StackedHN} proposed Stacked Hourglass Networks, employing a repeated downsampling and upsampling architecture to aggregate features across scales, improving the precision in localizing key points. Xiao~\etal~\cite{Xiao2018SimpleBF} further improves performance by designing a simple architecture that stacks transposed convolution layers in ResNet~\cite{He2015DeepRL} to produce high-resolution heat maps.
Sun \etal~\cite{Sun2019DeepHR} proposed HRNet, a network designed to maintain high-resolution and multi-scale representations throughout the entire process to achieve spatially accurate heat map, significantly improving accuracy. Inspired by HRNet's success, Yu~\etal\cite{Yu2021LiteHRNetAL} proposed Lite-HRNet, a light-weight version that utilized conditional channel weighting blocks to exchange information between different channels and resolutions. 

While CNN-based approaches have led to remarkable advancements in human pose estimation, they come with inherent limitations, particularly when compared to the capabilities of recent developments leveraging Vision Transformers (ViTs). One notable disadvantage is their inherent local receptive fields, which can sometimes limit their ability to capture long-range dependencies and the global context as effectively. This limitation can affect the accuracy of human pose estimation, especially in complex scenes where understanding the broader context is important.

\subsection{Vision Transformer Based Approaches}

\myparagraph{Vision Transformer and Its Challenges} The Vision Transformer (ViT)~\cite{Dosovitskiy2021AnII} is a state-of-the-art architecture that has gained increasing attention in the computer vision field. ViT processes an image as a sequence of $16\times 16$ patches, each one represented as a token vector. These patch embeddings are fed to a transformer encoder, which captures global relationships among all patches and outputs a global representation of the image. This representation is then fed to a simple head to make predictions and has demonstrated state-of-the-art performance in image classification.

Despite their effectiveness, ViTs use global self-attention to capture long-range dependencies in images, leading to a quadratic computational complexity with respect to the number of tokens.
Furthermore, for ViTs to achieve state-of-the-art performance, they need to be trained on large-scale datasets such as ImageNet-22K~\cite{Deng2009ImageNetAL} and JFT300M~\cite{Sun2017RevisitingUE}, which typically requires massive computational resources.  

\myparagraph{Addressing Computational Complexity in ViTs} To address this issue, several works have proposed various methods, including local attention mechanisms~\cite{Li2021LocalViTBL, Liu2021SwinTH}, sparse attention mechanisms~\cite{Roy2021EfficientCS, Child2019GeneratingLS}, and data-efficient image transformers~\cite{Touvron2021TrainingDI}. Additionally, numerous works focus on reducing the number of tokens that need to be processed by ViTs, thereby lowering their computational cost. 

For example, Token Learner~\cite{Ryoo2021TokenLearnerWC} is one approach that aims to merge and reduce the input tokens into a small set of important learned tokens. Token Pooling~\cite{Marin2021TokenPI} clusters the tokens and down-samples them, whereas DynamicViT~\cite{Rao2021DynamicViTEV} introduces a token scoring network to identify and remove redundant tokens. 
Adaptive Token Sampler~\cite{Fayyaz2022AdaptiveTS} adaptively down-samples input tokens by assigning significance scores to every token based on the attention weights of the class token in ViT. Similarly, EViT~\cite{Liang2022NotAP} determines tokens' importance scores via attention weights.  

Although these techniques successfully reduce the computational complexity of ViTs in classification tasks, the additional pooling and scoring network can introduce extra computational overhead. Moreover, the extension of these approaches to dense prediction tasks, such as human pose estimation, remains an open question.

\myparagraph{Transformer-based Human Pose Estimation} One of the earliest transformer-based approaches to human pose estimation, known as TransPose, was proposed by Yang~\etal~\cite{Yang2021TransPoseKL}. This vanilla transformer network estimates 2D poses from images via features extracted by CNN encoders and employs single-head self-attention within the transformer to model the long-range dependencies. 

Li~\etal~\cite{Li2021TokenPoseLK} proposed TokenPose, another transformer-based model that leverages CNN features and incorporates learnable joint tokens to explicitly embed each joint. Visual tokens and joint tokens are then fed to a standard transformer with global self-attention. To obtain the predicted heat maps, the joint tokens are mapped to a 2D feature vector by linear projection. Although visual tokens are simultaneously updated by the transformer in all layers, they are ignored during heat-map decoding, resulting in sub-optimal performance. 

Xu \etal\cite{Xu2022ViTPoseSV} introduced ViTPose, a ViT-based approach that uses a shared ViT encoder trained on multiple datasets to improve performance. However, it retains ViT's inherent limitations of quadratic computational complexity and fixed patch scale. 
To mitigate the computational complexity, \methodvit\ utilizes learnable tokens that explicitly learn joint embeddings to select patches with the most important information about body joints. 
In addressing multi-dataset training, ViTPose uses a shared encoder, whereas our unified training scheme advances upon this by enabling shared encoder and decoder training. This not only simplifies the training process but also enhances the performance.

\myparagraph{Multi-Scale Representation in ViTs} Unlike most CNN-based architectures~\cite{He2015DeepRL, Yu2021LiteHRNetAL}, vanilla ViT-based methods maintain patches of the same size in all layers, generating a fixed-scale representation. This is not ideal for dense prediction tasks such as human pose estimation, where people may appear at different scales in an image. 

Multi-scale ViTs~\cite{Wang2021PyramidVT, Liu2021SwinTH, Dong2022CSWinTA} address ViT's fixed-scale and quadratic computational complexity issues by constructing a hierarchical representation and limiting self-attention to a local window, respectively. To get a multi-scale image representation, multi-scale ViTs construct hierarchical representations by starting from small-sized patches and gradually merging neighboring patches. For example, Liu~\etal~\cite{Liu2021SwinTH} introduced Swin Transformer, a hierarchical ViT whose representation is computed with shifted windows, as a general-purpose backbone for computer vision. Dong~\etal~\cite{Dong2022CSWinTA} improved the performance by using cross-shaped window attention and locally-enhanced positional encoding.

Following these, Yuan~\etal~\cite{Yuan2021HRFormerHT} proposed HRFormer, a transformer-based pose estimation network that adopts HRNet's~\cite{Sun2019DeepHR} multi-resolution parallel design along with local-window self-attention and depth-wise convolution. Similarly, Xiong~\etal~\cite{Xiong2022SwinPoseST} uses a pre-trained Swin Transformer to extract features and utilize a feature pyramid structure for pose estimation. However, the utilization of local window self-attention restricts the network's modeling capability compared to global self-attention.
\method, similarly, uses a multiscale encoder with local window attention and enhances it by incorporating global context through the Joint Aware Global-Local (JAGL) attention mechanism. 
This is achieved by efficiently capturing the global context leveraging a small number of learnable joint tokens and propagating it back to the patch tokens. 

%%%%%%%%%%%%%%%% EViTPose %%%%%%%%%%%%%%%%
\section{\methodvit: ViT-Based Human Pose Estimation with Patch Selection}
\label{sec:method}

\begin{figure*}[!htbp]
\centering
  \subfloat[\label{fig:vitpose}]{%
    \includegraphics[width=0.99\linewidth]{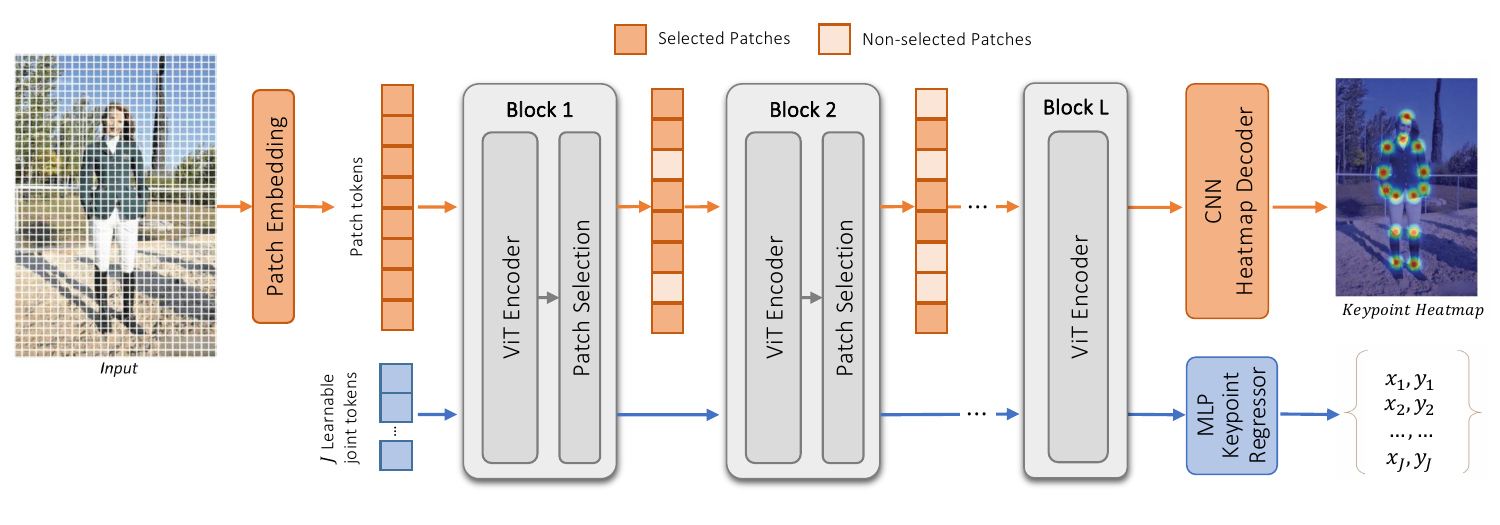}
  }
  \caption{\textbf{Overall architecture of
  \methodvit: ViT-based human pose estimation method with patch selection
  -- } An image is passed through a patch embedding layer to obtain patches of size $16\times 16$. These patches, along with $J$ learnable joint tokens, are processed by a ViT with $L$ transformer blocks. 
  Utilizing the joint tokens, the patch selection module progressively selects patches that more likely contain the most important information about body joints across all blocks except the last one. The non-selected patches are not processed by the subsequent blocks but are utilized in the heat-map decoder.
  The output of the final ViT block is then used by a CNN-based heat-map decoder to estimate the heat map of $J$ joints, while a simple MLP joint regressor estimates joints directly from the joint tokens. 
  }
  \label{fig:vitpose_skel}
  
\end{figure*}

In this section we describe \methodvit, an efficient vision transformer for human pose estimation that uses learnable tokens to select a small number of patches to be processed.
The overall architecture of \methodvit\ is shown in Figure~\ref{fig:vitpose}.

\subsection{ViT Encoding}
Given an input image $\mathbf{X} \in \mathbb{R}^{H \times W \times 3}$, where $(H, W)$ is the resolution of the image, the task is to find a mapping from $\mathbf{X}$ to the target 2D joint coordinates $Y\in \mathbb{R}^{J\times 2}$, where $J$ is the number of body joints for each person in the image. We first divide $\mathbf{X}$ into patches of size $16\times 16$, resulting in a 
set of patch tokens $\mathbf{P}\in \mathbb{R}^{N\times C}$, 
where $N=\left\lceil \frac{H}{16} \times \frac{W}{16}\right\rceil$ and $C$ represents the channel dimension. We then include $J$ learnable joint tokens $\mathbf{J} \in \mathbb{R}^{J\times C}$, which explicitly embed each joint and are later used to regress the joint 2D coordinates in the image. Next, we concatenate patch and joint tokens to form a matrix of input tokens $\mathbf{T}\in \mathbb{R}^{(N+J)\times C}$. 
The concatenated tokens are fed to a standard ViT encoder with $L$ transformer blocks, each consisting of a multi-head self-attention (MSA) layer and a fully connected MLP layer. 
Specifically, in each self-attention layer  the output tokens $\mathbf{O}$ are computed as
\begin{align}
 \mathbf{O} &= \text{Attn}(\mathbf{Q},\mathbf{K},\mathbf{V}) = \mathcal{A} \mathbf{V}, \quad \quad \mathcal{A} = \text{Softmax}( \frac{\mathbf{Q} \mathbf{K}^\top}{\sqrt{C}}),
\end{align}
where $\mathbf{Q}\in \mathbb{R}^{(N+J)\times C}$, $\mathbf{K}\in \mathbb{R}^{(N+J)\times C}$ and $\mathbf{V} \in \mathbb{R}^{(N+J)\times C}$ are queries, keys, and values, respectively, which are computed from the input tokens $\mathbf{T}$ as in the standard ViT~\cite{Dosovitskiy2021AnII}, 
and $\mathcal{A}\in \mathbb{R}^{(N+J)\times (N+J)}$ is the attention matrix.

Given the final feature map of the patches produced by ViT, 
we use a classical decoder with two deconvolution blocks, each with a deconvolution layer, batch normalization, and ReLU activation to estimate the heat map of $J$ joints. Similarly,  a LayerNorm layer followed by a fully connected MLP layer is used to directly regress the $J$ joint coordinates from the joint tokens. In this way, the joint tokens are enforced to learn the important joint-level information to be able to successfully regress the joint 2D coordinates.

\subsection{Improving Efficiency via Patch Selection}
Although ViT can model long-range dependencies and is able to generate a global representation of the image, the computational complexity increases quadratically with the number of tokens. However, not all patches in an image contribute equally to the human pose estimation task. Recent research~\cite{Yang2021TransPoseKL} indicates that the long-range dependencies between predicted joints are mostly restricted to the body part regions. Therefore, computing MSA between every patch in the image is unnecessary as only a few patches are relevant to the body parts.
To this end, we propose to select a small number of relevant patches while discarding irrelevant and background patches without re-training the vision transformer. By selecting only the relevant patches, we can significantly reduce the computational complexity as shown in~\cite{Fayyaz2022AdaptiveTS, Rao2021DynamicViTEV, Meng2021AdaViTAV} for the classification task.

\subsubsection{Off-the-Shelf Pose Estimator Based Patch Selection} %Methods
One approach to selecting a small subset of relevant patches is to use an off-the-shelf lightweight pose estimation network to identify a small number of patches that are more likely to contain the joints. Two methods introduced in our previous work~\cite{EViTPose} follow this approach. 
The first method employs a breadth-first neighboring search algorithm to select body joint patches and their neighbors based on estimated pose predictions. Extending this, the second method selects patches formed by a skeleton of joints, aiming to identify body patches where lines formed by joint pairs intersect, utilizing Bresenham's algorithm~\cite{Bresenham1965AlgorithmFC}. 

While the aforementioned methods can remove irrelevant patches before they are processed by ViT and thus enhance its efficiency, their reliance on the accuracy of off-the-shelf pose estimators is a limitation. In this work, we present an alternative approach for automatically selecting body part patches via learnable joint tokens that enable the selection of relevant patches using their corresponding attention maps, as outlined in Section~\ref{sec:auto_patch_sel}.

% \subsection{Autonomous Patch Selection Method}
\subsubsection{Joint-Token-Based Patch Selection Method}
\label{sec:auto_patch_sel}
The limitation of the first two patch selection methods introduced in~\cite{EViTPose} is that they rely on the performance of the lightweight off-the-shelf pose estimator. This limitation can become especially problematic when dealing with complex scenes, as the accuracy of the pose estimator is often compromised by occlusion, motion, or variations in camera perspective. As a consequence, this might result in the selection of irrelevant patches and removing important patches, leading to suboptimal performance.
Therefore, a more robust approach is required that can adapt to these challenging scenes without the need for an off-the-shelf pose estimator. 
To overcome this limitation, we propose an approach that involves two key strategies: (i) selecting patches that more likely contain the most important information about body joints using a small number of learnable joint tokens that effectively capture the essential features of body joints, and (ii) refining the representation of non-selected patches by leveraging the global information extracted by the joint tokens to update non-selected patch tokens in a computationally efficient manner before their exclusion from further processing in ViT. Since these non-selected patches will later be utilized in the heat-map decoding stage, having a more refined representation is beneficial. 

\myparagraph{Selecting most informative patches via learnable joint tokens}
We select the most important patches using the learnable joint tokens,
which serve as a powerful feature representation for distinguishing the relevant body part patches.
Specifically, we aim to determine the importance of each patch in relation to the joint tokens, thereby enabling us to select the most informative body part patches. To achieve this, we harness the attention matrix similar to~\cite{Liang2022NotAP, Fayyaz2022AdaptiveTS, Goyal2020PoWERBERTAB}, as the values in $\mathcal{A}$ represent the weight of contribution of input tokens to output tokens. For example, 
$\mathcal{A}_{N+1:N+J,1}$ denotes the attention weights from the first patch token to the output tokens ranging from the $(N+1)^{\text{th}}$ to the $(N+J)^{\text{th}}$ positions,
which correspond to the $J$ joint tokens. Thus, we can calculate the average contribution weight of a patch token $l$ to the $J$ joint tokens as follows:
\begin{equation}
{\mathcal{W}}_l = \frac{1}{J} \sum_{j=1}^J \mathcal{A}_{N+j, l}.
\end{equation}
Following~\cite{Fayyaz2022AdaptiveTS}, we take the norm of the value of token $l$, $\mathbf{V}_l$, into account for calculating the importance score. Thus, the importance score of the patch token $l$ is:
\begin{equation}
\mathcal{I}_l = \frac{{\mathcal{W}}_l\times {\|\mathbf{V}_l\|}}{ \sum_{k=1}^N {\mathcal{W}}_k \times \|\mathbf{V}_k\|},
\end{equation}
where $l,k \in \{1,\dots,N\}$.
Once the importance scores of each patch token have been computed, we select the $L \ll\ N$ patch tokens with the highest scores for further processing. 

\myparagraph{Pruning attention matrix} Our subsequent step involves pruning the attention matrix $\mathcal{A}\in \mathbb{R}^{(N+J)\times (N+J)}$ by selecting the rows that correspond to the selected $L$ patch tokens and $J$ joint tokens, designated as $\mathcal{A}^s \in \mathbb{R}^{(L+J)\times (L+J)}$. Then the output tokens $\mathbf{O}^s\in \mathbb{R}^{(L+J)\times C}$ are calculated as follows:
\begin{equation}
\mathbf{O}^s = \mathcal{A}^s \mathbf{V}^s,
\end{equation}
where $\mathbf{V}^s$ corresponds to the values of the selected tokens. These output tokens are then passed as input for the next blocks. 

\myparagraph{Refining non-selected patches via joint tokens} Although only $\mathbf{O}^s$ will be processed in the next blocks of ViT, the non-selected patch tokens will still be used during the heat-map decoding.
Therefore, it is important to have a refined representation of the non-selected patch tokens before they are excluded from further processing in the next blocks. Thus, we propose an efficient method that updates these tokens using the joint tokens only. This approach is motivated by the fact that joint tokens learn global information and therefore can be used to update the patch tokens in a computationally efficient manner without the need to compute contributions from all tokens, which can be computationally expensive. We start by selecting the rows of the attention matrix $\mathcal{A}$ that correspond to the non-selected patch tokens and the columns that correspond to the joint tokens, resulting in a sub-matrix $\mathcal{A}^o \in \mathbb{R}^{(N-L) \times J}$. We then update the non-selected patch tokens using the joint tokens as follows:
\begin{equation}
\mathbf{O}^o = \mathcal{A}^o \mathbf{V}^j,
\end{equation}
where $\mathbf{V}^j \in \mathbb{R}^{J \times C}$ corresponds to the values of the joint tokens. 

%%%%%%%%% END OF EVITPOSE %%%%%%%%%%%%%

%%%%%%%%% UniTransPose %%%%%%%%%%%%%%%%
\begin{figure*}[!htbp]
  \centering
  \subfloat{\includegraphics[width=0.7\textwidth]{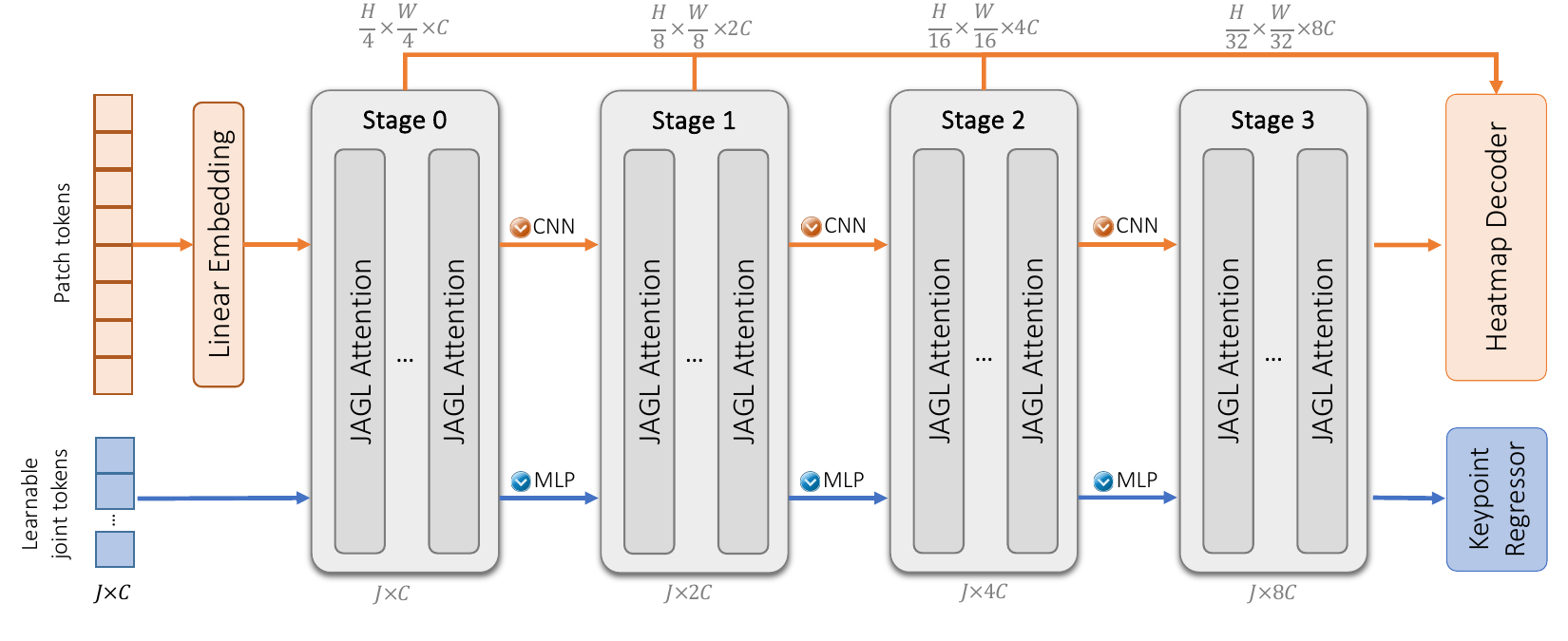}}
  \hspace{0.01\textwidth}
  \color{gray}
  \rule[1.5ex]{0.05pt}{28ex} % \vline
  \color{black}
  \hspace{0.01\textwidth}
  \subfloat{\includegraphics[width=0.261\textwidth]{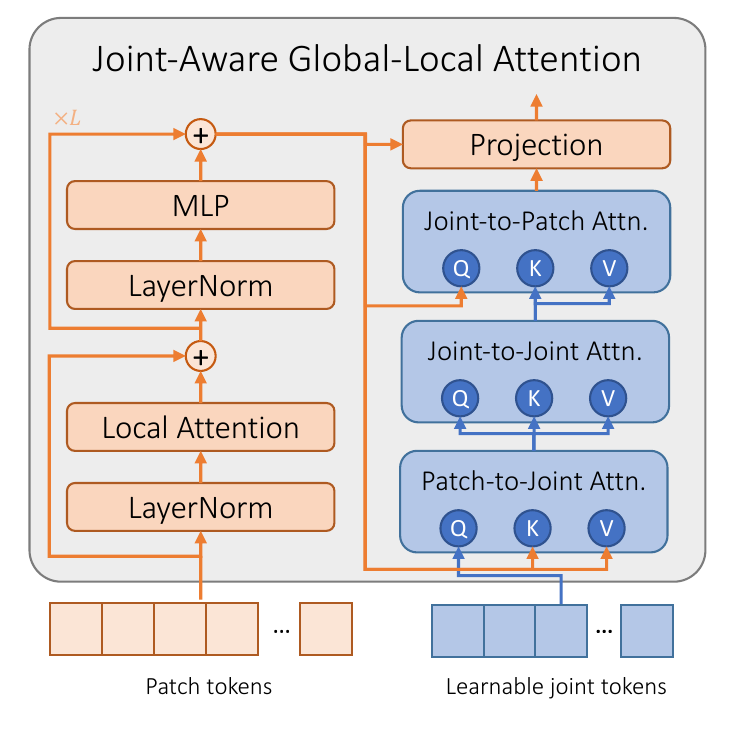}}
  \caption{\textbf{Overall architecture of \method, a multi-scale vision transformer based human pose estimation network -- } An input image $X \in \mathbb{R}^{H\times W\times 3}$ is fed into a patch embedding layer that divides the image into patches of size $4\times 4$. A linear embedding layer then projects the patch tokens to a $C-$dimensional vector.
  % , resulting in a feature map $F \in \mathbb{R}^{\frac{H}{4} \times \frac{W}{4} \times C}$. 
  The patch tokens along with joint tokens are processed by four stages. Each stage comprises Joint-Aware Global Local (JAGL) attention blocks, which consist of local patch-to-patch attention followed by global patch-to-joint, joint-to-joint, and joint-to-patch attention.  
  %At each adjacent stage, 
  A convolution layer ($3 \times 3$, stride $2$) is applied between stages to reduce the spatial resolution of the patch tokens and generate a hierarchical feature map. This operation also doubles the patch tokens' channel dimension. Consequently, to maintain consistency, a linear embedding layer is used to double the channel dimension of the joint tokens.
  The output of each stage is then passed to a CNN-based decoder to estimate the heat map of the $J$ joints. Meanwhile, the key-point regressor uses the joint tokens to directly estimate the $(x,y)$ locations of the $J$ joints.}
  \label{fig:mspose}
\end{figure*}

\section{\method: Multiscale Transformer-based Human Pose Estimation}
In this section, we describe our second approach, \method: a multi-scale vision transformer for human pose estimation 
that addresses the quadratic computational complexity and fixed patch scales of ViTs by restricting the self-attention computation to a local window and constructing hierarchical feature maps, respectively. It further enhances the modeling capability of the model by capturing the global context via the Joint-Aware Global-Local attention, which will be detailed further in Section~\ref{sec:jagl}.
We present the overall architecture of our model, \method, in Fig. \ref{fig:mspose}.

\subsection{Hierarchical Architecture}
Similar to several multi-scale ViTs~\cite{Wu2021CvTIC, Wang2021PyramidVT, Liu2021SwinTH, Dong2022CSWinTA}, \method\ employs a ViT encoder that constructs a hierarchical representation by starting from small-sized patches and gradually merging neighboring patches in deeper layers, thus it has the flexibility to model various scales while maintaining a linear computational complexity. Following~\cite{Dong2022CSWinTA}, \method\ utilizes a cross-shaped local window for simultaneously computing self-attention in both horizontal and vertical directions. Additionally, it introduces a global attention mechanism that efficiently captures the global contextual information across the entire image.

As shown in Figure~\ref{fig:mspose}, given an input image $X \in \mathbb{R}^{H\times W\times 3}$, the image is embedded into $4\times 4$ patches, resulting in patch tokens, $\mathbf{P}\in \mathbb{R}^{N \times C}$, where $N=\left\lceil \frac{H}{4} \times \frac{W}{4} \right\rceil$ and $C$ is the channel dimension, using 
Convolutional Token Embedding
from~\cite{Wu2021CvTIC}. Additionally, we include $J$ learnable joint tokens $\mathbf{J} \in \mathbb{R}^{J\times C}$, corresponding to each joint in an image. Both patch and joint tokens are then processed by 4 stages each containing a different number of Transformer blocks. 
A convolutional layer ($3 \times 3$, stride $2$) is applied at each subsequent stage to reduce the spatial resolution of the patch tokens and generate a hierarchical feature map. This operation also doubles the channel dimension of the patch tokens. As a result, a linear embedding layer is used to double the channel dimension of the joint tokens to maintain consistency.
Therefore, the output of the four stages is $F_0 \in \mathbb{R}^{\frac{H}{4} \times \frac{W}{4} \times C }$, $F_1 \in \mathbb{R}^{\frac{H}{8} \times \frac{W}{8} \times 2C }$, $F_2 \in \mathbb{R}^{\frac{H}{16} \times \frac{W}{16} \times 4C }$, and $F_3 \in \mathbb{R}^{\frac{H}{32} \times \frac{W}{32} \times 8C }$, respectively.

\subsection{Joint Aware Global-Local (JAGL) Attention}
\label{sec:jagl}

As discussed before, the standard Multi-head Self-Attention (MSA) mechanism used by ViT is a powerful technique for capturing long-range interactions among all patches in an image. However, 
MSA suffers from a quadratic computational complexity with respect to the number of tokens. Local window-based self-attention addresses this challenge by limiting the computation of attention to a small window of patches, which reduces the computational complexity to linear with respect to the number of patches. However, this also reduces the expressivity of the model.
The challenge is, hence, to design an attention mechanism that both captures global information and is computationally efficient.

To address this, we propose to combine both local and global attention. Specifically, we propose to compute local attention among patch tokens to allow for better scalability, and to compute global attention with the joint tokens, since the number of patch tokens is much greater than the number of joint tokens. The joint tokens are trained to predict joint locations and hence capture global information. Thus, the joint tokens serve as a bottleneck to efficiently share this global information with the patch tokens without the need for the computationally expensive global self-attention between the patches.  We achieve this by using a Joint Aware Global-Local (JAGL) attention mechanism, which combines local patch-to-patch attention and global patch-to-joint, joint-to-joint, and joint-to-patch attention, as described next.

\subsubsection{Local Patch-to-Patch Attention}
We use the cross-window self-attention approach~\cite{Dong2022CSWinTA}, which enhances local self-attention by considering the interactions in a cross-shaped window (CW). Specifically, we divide the $M$ heads into two groups
\begin{align}
    \text{CW-Head}^{m} = \begin{cases}
      \text{H-Attn}^{m}(\mathbf{P}, w) & \text{if $m \in [ 1,\frac{M}{2}]$ }\\
      \text{V-Attn}^{m}(\mathbf{P}, w) &  \text{if $m \in (\frac{M}{2},M]$} \\
    \end{cases},
\end{align}
where $m$ is the head index, and H-Attn and V-Attn denote horizontal and vertical stripe self-attention with window size $w$, respectively. Finally, we combine the outputs of both groups as follows:
\begin{align}
\hat{\mathbf{P}} =  \text{Local-Attn}(\mathbf{P}) = \text{Concat}_{\{m\}}(\text{CW-Head}^{m})\mathbf{W}_L,
\end{align}
where $\mathbf{W}_L \in \mathbb{R}^{C\times C}$ is a matrix that projects the local-attention result into the target output dimension. 

\subsubsection{Global Patch-to-Joint Attention} 
The global attention layer first computes the cross-attention between joint tokens and patch tokens updated via local attention. Specifically, given the updated patch tokens, $\hat{\mathbf{P}}$, and the joint tokens represented by a matrix $\mathbf{J}$, we extract the joint-level global context $\hat{\mathbf{J}}$ from the patch tokens using the following cross-attention mechanism with the joint tokens as the queries and the patch tokens as  keys and values:
\begin{align}
    \hat{\mathbf{J}} = \text{Concat}_{\{m\}}(\text{Attn}(\mathbf{W}_Q^m \mathbf{J}^m, \mathbf{W}_K^m \hat{\mathbf{P}}^m, \mathbf{W}_V^m \hat{\mathbf{P}}^m)), 
\end{align}
where $\mathbf{W}_Q$, $\mathbf{W}_K$ and $\mathbf{W}_V$ denote the projection matrices for the queries, keys, and values, respectively. 

\subsubsection{Global Joint-to-Joint Attention} We then perform global self-attention among the joint tokens as follows:
\begin{align}
    \Tilde{\mathbf{J}} = \text{Concat}_{\{m\}}(\text{Attn}(\hat{\mathbf{W}}_Q^m \hat{\mathbf{J}}^m, \hat{\mathbf{W}}_K^m \hat{\mathbf{J}}^m, \hat{\mathbf{W}}_V^m \hat{\mathbf{J}}^m)).
\end{align}
where $\hat{\mathbf{W}}_Q$, $\hat{\mathbf{W}}_K$, and $\hat{\mathbf{W}}_V$ denote the projection matrices for the queries, keys, and values, respectively. 

\subsubsection{Global Joint-to-Patch Attention} Next, we compute the cross-attention between patch tokens and updated joint-tokens by using the patch tokens as queries and the joint-tokens as both keys and values:
\begin{align}
    \Tilde{\mathbf{P}} &= \text{Concat}_{\{m\}}(\text{Attn}(\Tilde{\mathbf{W}}_Q^m \hat{\mathbf{P}}, \Tilde{\mathbf{W}}_K^m \Tilde{\mathbf{J}}^m, \Tilde{\mathbf{W}}_V^m \Tilde{\mathbf{J}}^m)),
\end{align}
where $\tilde{\mathbf{W}}_Q$, $\tilde{\mathbf{W}}_K$ and $\tilde{\mathbf{W}}_V$ denote the projection matrices for the queries, keys, and values, respectively. Finally, we concatenate both updated patch representations as follows:
\begin{align}
    \mathbf{P} &= \text{Concat}({\Tilde{\mathbf{P}}, \hat{\mathbf{P}}}) \mathbf{W}_{P},
\end{align}
where $\mathbf{W}_{P}$ is a projection matrix that projects concatenated features to the same dimension as patch tokens $\mathbf{P}$. 

Therefore, JAGL efficiently captures the global context utilizing a limited number of learnable joint tokens to gather and then disseminate it back to the patch tokens, avoiding the computationally expensive global self-attention among the significantly larger number of patch tokens. 

\section{Hybrid Key-Point Decoding and Unified Skeletal Training}
As previously discussed, the two most common methods for decoding key-points are heat-map decoding and key-point regression. While key-point regression is more efficient than heat-map decoding, it is also less accurate and less robust~\cite{Yu2020heatmapRV,Tompson2014JointTO,Nibali2018NumericalCR}. Here, we propose to improve the efficiency of classical heat-map decoders with an efficient sub-pixel CNN-based heat-map decoder. In addition, we train our models with both decoding options, 
so that during inference users can choose either the key-point regressor for efficiency or the heat-map decoder for accuracy and robustness.

\subsection{Efficient Sub-Pixel CNN-Based Heat Map Head}
In heat-map decoding, the encoder feature maps are fed to a decoder that produces Gaussian heat maps for each joint. In the case of multi-scale encoders, hierarchical feature maps are fed into decoders. Most previous approaches rely on classical CNN-based decoders~\cite{Xiao2018SimpleBF, Xu2022ViTPoseSV}. For multi-scale encoders, hierarchical feature maps are often processed by decoders such as Feature Pyramid Networks~\cite{Lin2016FeaturePN, He2014SpatialPP}. However, this approach typically involves upsampling the low-resolution feature maps to the target resolution using bicubic interpolation before employing a convolutional network, which multiplies the number of parameters and the amount of computational power required for  training by the square of the desired up-sample scale. 
% TODO: 
Some of the recent works address this by using a simple decoder that only uses the last feature map~\cite{Sun2019DeepHR, Yuan2021HRFormerHT}. However, this could limit the model's ability to handle humans on multiple scales.

To address this issue, we propose a Pixel-Shuffle-based decoder that employs a convolutional network in the lower-resolution image, instead of the desired output resolution, and upsamples it using the Pixel-Shuffle operation. 
Pixel Shuffle~\cite{Shi2016RealTimeSI} was originally proposed for super-resolution tasks and has proven to be an efficient method. It preserves information and achieves the same result as regular transpose convolution, but it requires fewer channels in the higher-resolution feature map.
Specifically, given a low-resolution feature map $\mathbf{F}^{LR}\in \mathbb{R}^{H\times W\times C \cdot r^2}$, where $r$ is an upsampling factor, we apply 2 layers of convolutions with a kernel size of $1\times 1$ followed by two layers of convolutions with a kernel size of $3 \times 3$ directly on the low-resolution feature map, which greatly reduces the computational complexity. We then apply the pixel shuffle operation, $F^{HR} = \mathcal{PS}(F^{LR}, r)$, to the pixels of the processed low-resolution feature map to obtain a high-resolution feature map $F^{HR} \in \mathbb{R}^{H\cdot r \times W\cdot r \times C}$ as:
\begin{align}
    %\mathcal{PS}(F^{LR}, r)_{(x,y,c)} &= 
    \mathbf{F}^{HR}_{(x,y,c)} = 
    \mathbf{F}^{LR}_{(\lfloor \frac{x}{r} \rfloor, \lfloor \frac{y}{r} \rfloor, C\cdot r\cdot \text{mod}(y,r) + C\cdot \text{mod}(x,r) + c )},  
\end{align}
where $x, y$ are the output pixel coordinates in the high-resolution space and $c$ is the channel index. After the feature maps are upsampled to the desired output resolution, we concatenate them and apply two convolutional layers with kernel size $3\times 3$ and $1\times 1$, respectively.

It is also possible to further increase the efficiency of our model by utilizing only the last feature map, instead of the hierarchical feature maps, similar to what the recent methods do. However, this comes at the expense of reduced accuracy. This approach may be appropriate for certain use cases where speed is more important than accuracy.

\subsection{Simple Key-point Regression Head}
Key-point regression involves predicting the exact coordinates of joints in an image, rather than predicting a heat map that indicates the likelihood of a joint being present at each location (as in heat-map decoding). To do this, a network must predict the $x$ and $y$ coordinates of each joint separately. In our method, one simple approach is to use a LayerNorm (LN) layer followed by a fully connected MLP layer to directly regress the $J$ joint coordinates, $\mathbf{\hat{Y}} \in \mathbb{R}^{J\times 2}$, given the joint tokens, $\mathbf{J} \in \mathbb{R}^{J\times C}$, as follows:
\begin{align}
    \mathbf{\hat{Y}} = \text{MLP}(\text{LN}(\mathbf{J})).
\end{align}

\subsection{Unified Skeletal Representation and Training}
\begin{figure*}[htbp]
  \centering
  % \vspace{-5mm}
  \subfloat[COCO, \\ \qquad {\color{white}....} OCHuman]{\includegraphics[width=0.146\textwidth]{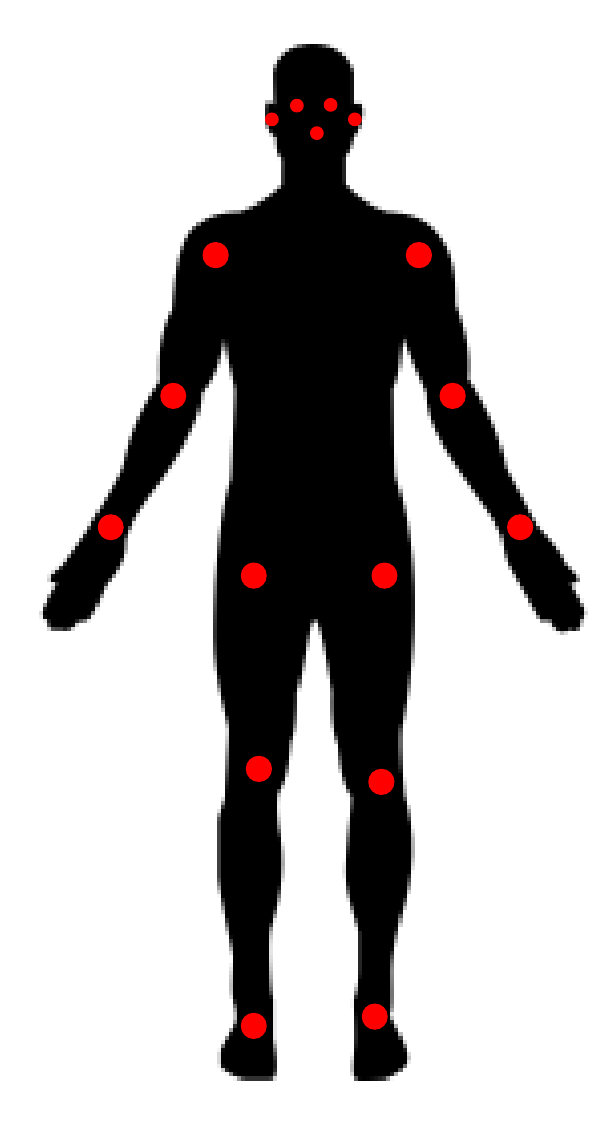}}
  \hspace{1.5mm}
  \subfloat[JRDB]{\includegraphics[width=0.146\textwidth]{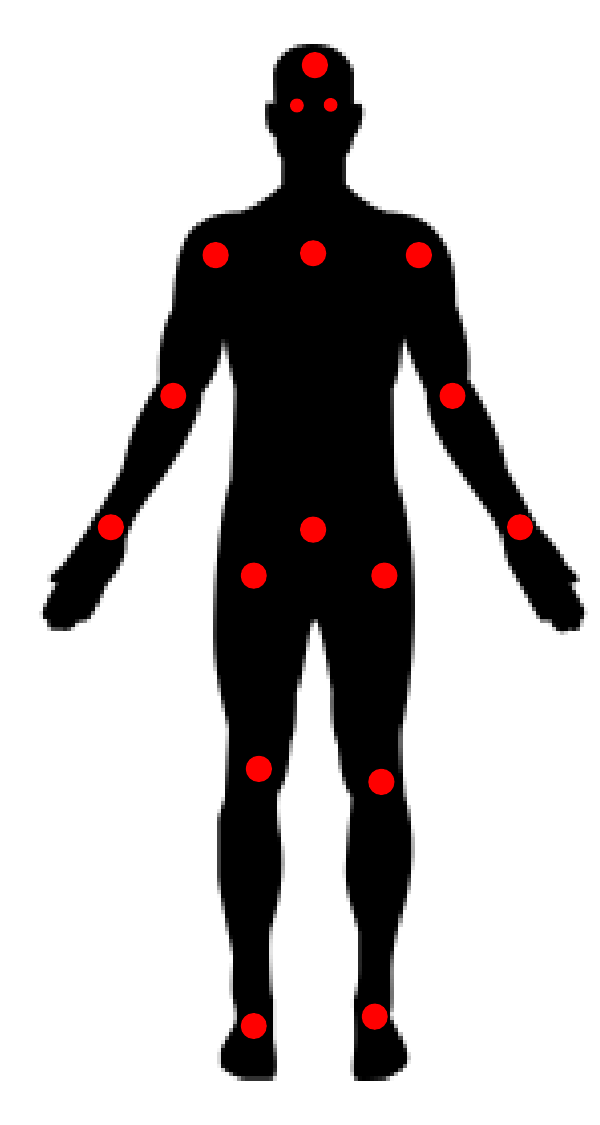}}
  \hspace{1.5mm}
  \subfloat[MPII]{\includegraphics[width=0.146\textwidth]{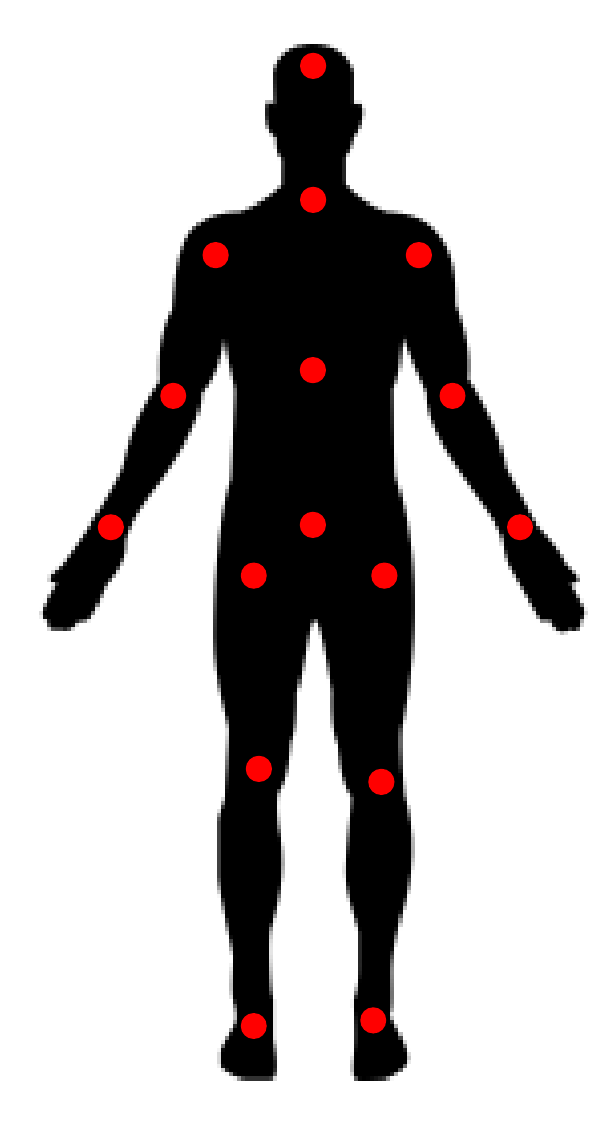}}
  \hspace{1.5mm}
  \subfloat[AIChallenger, \\ \qquad {\color{white}....} CrowdPose]{\includegraphics[width=0.146\textwidth]{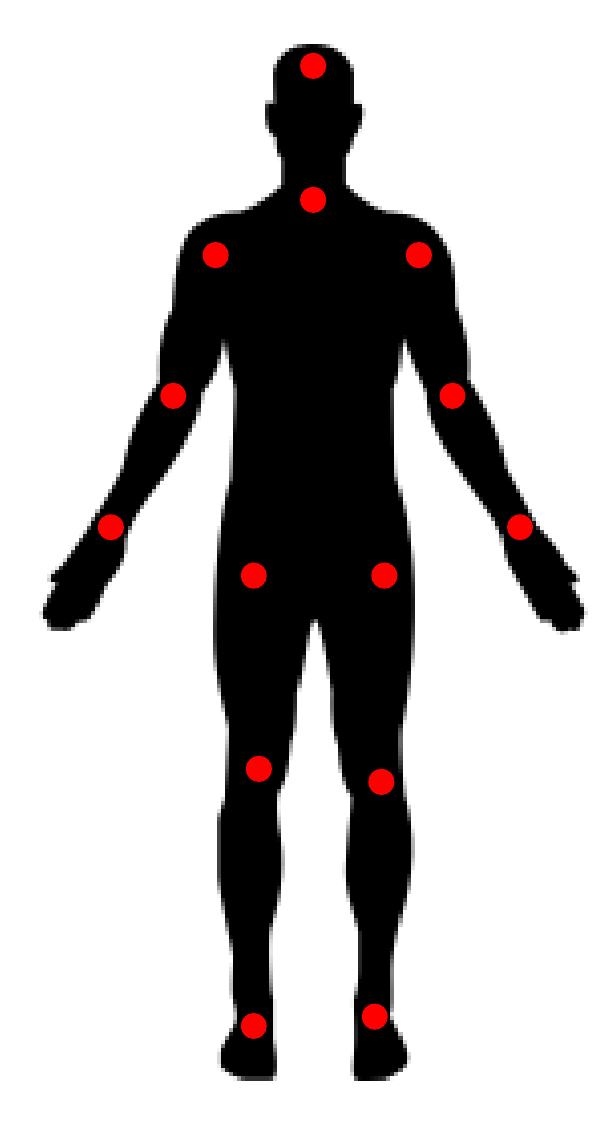}}
  \hspace{1.5mm}
  \hspace{0.01\textwidth}
  \color{gray}
  \rule[1.5ex]{0.05pt}{28ex} % \vline
  \color{black}
  \hspace{0.01\textwidth}
  \subfloat[Unified]{\includegraphics[width=0.146\textwidth]{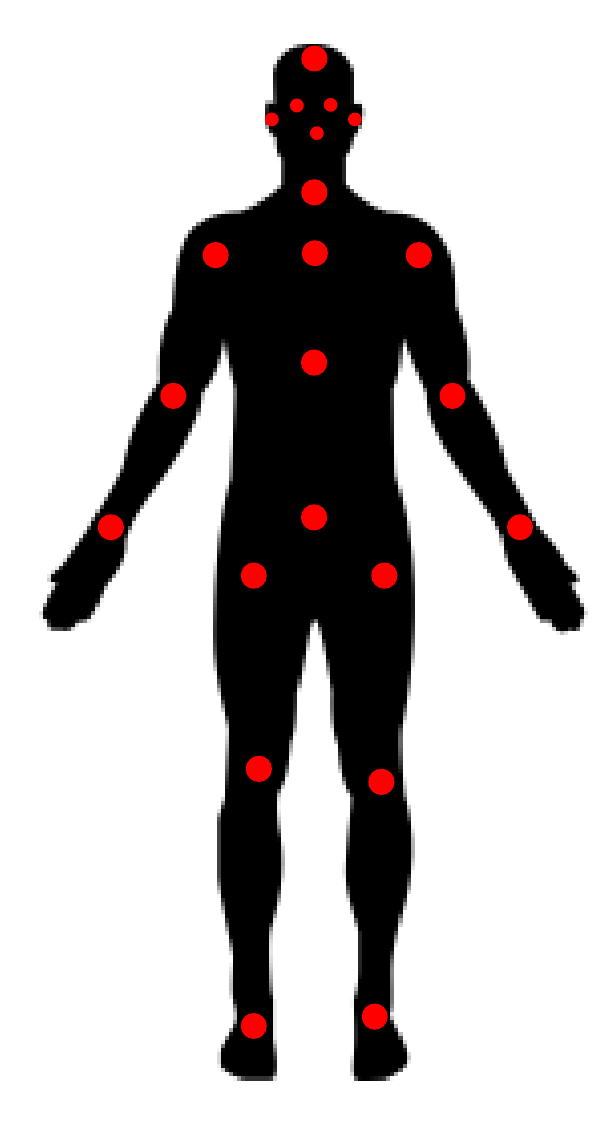}}
  % \vspace{-2mm}
  \caption{\textbf{Distinct annotation styles across multiple benchmarks.} (a) COCO and OCHuman share a common 17-joint skeleton. (b) JRDB uses the same number of joints but differs in locations. (c) MPII employs 16 joints. (d) AIChallenger and CrowdPose use 14 joints. (e) The proposed Unified skeleton comprises all joints present in the various benchmarks.}
  \label{fig:skels}
  % \vspace{-3mm}
\end{figure*}

Vision transformers have been shown to be powerful models for numerous computer vision tasks. However, they require large-scale datasets to perform well~\cite{Dosovitskiy2021AnII, Raghu2021DoVT}. Many existing training datasets could be used to help our model generalize to a wide range of human pose variations. However, these datasets can have different numbers of joints and annotation styles (see Fig~\ref{fig:skels} (a-d) and Sec~\ref{sec:datasets}), which makes it challenging to train one model on all datasets. 
% ViTPose~\cite{Xu2022ViTPoseSV} uses a shared encoder and a separate decoder for each dataset to improve the generatlizability 

To address this issue, we propose a unified skeletal representation that includes all joints from all datasets (see Fig~\ref{fig:skels} (e)). However, training on all joints requires defining a loss that can handle a number of predicted joints that is larger than the number of ground truth joints. To handle such variations, we use a weighted $L_1$ and $L_2$ loss function for joint coordinates and heat-map encoding, respectively:
\begin{align}
    {L}_1(\mathbf{\hat{Y}_r}, \mathbf{Y}_r) = \frac{1}{J} \sum_{j=1}^J w_j \|\mathbf{\hat{Y}}_r^j - \mathbf{Y}_r^j\|_1, \\
    {L}_2(\mathbf{\hat{Y}_h}, \mathbf{Y}_h) = \frac{1}{J} \sum_{j=1}^J w_j \|\mathbf{\hat{Y}}_h^j - \mathbf{Y}_h^j\|_2^2, 
\end{align}
where $\mathbf{\hat{Y}}_r$ and $\mathbf{Y}_r$ represent the predicted and ground-truth 2D joint coordinates, respectively, $\mathbf{\hat{Y}}_h$ and $\mathbf{Y}_h$ are the corresponding  Gaussian heat maps, $J$ is the number of joints of the unified skeletal representation, and $w_j\in[0,1]$ is the weight assigned to joint $j$. We assign a weight $w_j=0$ to joints that are not in the ground truth and a weight $w_j \in (0,1]$ to joints in the ground truth depending on the importance of the joint as provided by the benchmarks. 

Our models are trained using a weighted combination (with weight $\lambda>0$) of a heat-map decoding loss and a key-point regression loss, i.e.:
\begin{align}
    \mathcal{L} = {L}_2(\mathbf{\hat{Y}}_h, \mathbf{Y}_h) + \lambda \cdot \text{Smooth}_{{L}_1}(\mathbf{\hat{Y}}_r, \mathbf{Y}_r).
    % {L}_F(H_P, H_G) = \frac{1}{K} \sum_{k=1}^K w_k (\frac{1}{H\times W} \|H_G - H_P\|_F^2), 
\end{align}
The heat-map decoding loss, ${L}_2(\mathbf{\hat{Y}}_h, \mathbf{Y}_h)$, compares the $J$ predicted heat maps $\hat{Y}_h$ to the $J$ ground-truth Gaussian heat maps $Y_h$ using the weighted $L_2$ loss summed over all pixels in each one of the $J$ heat maps. The key-point regression loss, $\text{Smooth}_{{L}_1}(\mathbf{\hat{Y}}_r, \mathbf{Y}_r)$, is a smooth $L_1$ loss that compares the predicted $x$ and $y$ coordinate values $\mathbf{\hat{Y}}_r$ to the ground-truth coordinates $\mathbf{Y}_r$. The smooth $L_1$ loss combines the $L_1$ and $L_2$ loss functions with the threshold $\frac{1}{\beta^2}$ switching from the ${L}_1$ to ${L}_2$ loss function for targets in the range $[0, \frac{1}{\beta^2}]$, i.e.:
\begin{align}
    \text{Smooth}_{{L}_1}(\hat{Y}_R, Y_R) = 
    \begin{cases}
     \frac{\beta^2}{2} \cdot {L}_2, & \text{if } {L}_1 \leq \frac{1}{\beta^2} \\
     {L}_1 - \frac{1}{2\beta^2}, &  \text{otherwise}.  \\
    \end{cases}
\end{align}

The proposed unified approach has several advantages. First, it allows our model to learn from any dataset while being adaptable to different joint numbers and annotation styles. Second, it simplifies the training process by eliminating the need for separate models trained on different datasets, which most prior works do. Third, it allows us to have more key points during inference, giving us more flexibility in selecting the joints of interest for a particular application. Overall, this approach significantly improves the performance of our proposed multi-scale transformer model on all the benchmarks by learning from a diverse set of datasets while being adaptable to varying annotation styles and joint numbers.

%%%%%%%%% END OF UNITRANSPOSE %%%%%%%%%%%%%

\section{Experiments}
\label{sec:experiment}

\subsection{Dataset details}
\label{sec:datasets}
We evaluate our methods on six common 2D pose estimation benchmarks: MS-COCO~\cite{Lin2014MicrosoftCC}, AI Challenger~\cite{Wu2017AIC}, JRDB-Pose~\cite{Vendrow2022JRDBPoseAL}, 
MPII~\cite{Andriluka20142DHP},
CrowdPose~\cite{Li2018CrowdPoseEC} and OCHuman~\cite{Zhang2019Pose2SegDF}. 

%AI Challenger
The first five datasets are used to train and test the proposed methods, and OCHuman is used to further test the models' performance in dealing with occluded people.

\myparagraph{AI Challenger} A large-scale dataset containing over 300,000 images with a total of 700,000 human poses, annotated with 14 joints. The images were collected from a variety of sources and exhibit significant variability in terms of pose, lighting conditions, and image quality.

%JRDB-Pose
\myparagraph{JRDB-Pose} This dataset contains a wide range of difficult scenarios, such as crowded indoor and outdoor locations with varying scales and occlusion types. It contains 57,687 panoramic frames captured by a social navigation robot, with a total of 636,000 poses annotated with 17 joints.

%MS-COCO
\myparagraph{MS-COCO} Another large-scale dataset containing more than 200,000 images in difficult and unpredictable conditions, with 250,000 person instances labeled with 17 joints.

%MPII
\myparagraph{MPII Human Pose Dataset} A popular dataset extracted from YouTube videos that includes around 25,000 images containing over 40,000 people annotated with 16 joints. 

%CrowdPose
\myparagraph{CrowdPose} A benchmark specifically designed to challenge human pose estimation models in crowded scenes, where multiple individuals are present in the same image. It comprises over 20,000 images, with annotations for more than 80,000 human instances, each labeled with 14 joints. 

%OCHuman
\myparagraph{OCHuman} A testing benchmark of 4,731 images containing 8,110 heavily occluded people labeled with 17 joints.

\subsection{Evaluation metrics}
On the MPII benchmark, we adopt the standard PCKh metric as our performance evaluation metric. PCKh is an accuracy metric that measures if the predicted key point and the true joint are within a certain distance threshold ($50\%$ of the head segment length). On the remaining benchmarks, we adopt standard average precision (AP) as our main performance evaluation metric. AP is calculated using Object Keypoint Similarity (OKS) averaged over multiple OKS values ($.50:.05:.95$). OKS is defined as:
\begin{equation}
OKS = \frac{ \sum_i \exp( - \frac{d_{i}^2}{2s^2j^2_{i}} ) \sigma(v_i>0) }{\sum_i \sigma(v_i>0)},
\end{equation}
where $d_i$ is the Euclidean distance between the detected key point and the corresponding ground truth, $v_i$ is the visibility flag of the ground truth, $s$ is the person scale, $\sigma$ is the per-key-point standard deviation, and $j_i$ is a per-joint constant that controls falloff. OKS measures how close the predicted key-point location is to the ground-truth key point. OKS has a value between 0 and 1: the closer the predicted key point is to the ground truth, the closer OKS is to 1. 

\subsection{Implementation details}

As previously mentioned, all our experiments employ the common top-down setting for human pose estimation, i.e., a person detector is used to detect person instances and the proposed methods are used to estimate the location of the joints of the detected instances in the image. The performance is then evaluated using Faster-RCNN~\cite{Ren2015FasterRT} detection results for the MS-COCO key-point validation set, with a detection AP of
56.4 for the person category, following~\cite{Xiao2018SimpleBF}. 
We follow most of the default training and evaluation settings of mmpose\footnote{https://github.com/open-mmlab/mmpose, Apache License 2.0}, except that we change the optimizer to AdamW~\cite{Loshchilov2019DecoupledWD}, a variant of Adam shown to be more effective for Transformers. AdamW decouples $L_2$ regularization and weight decay with a learning rate of $5e-6$ and uses UDP~\cite{Huang2020TheDI} as post-processing. We use $\lambda = 1e-2$ and $\beta=1$.

\subsection{Results}

Table \ref{tab:all} presents a comprehensive comparison of our methods (\methodvit\ and \method), recent CNN-based methods such as HRNet~\cite{Sun2019DeepHR}, and recent transformer-based methods such as HRFormer~\cite{Yuan2021HRFormerHT} and ViTPose~\cite{Xu2022ViTPoseSV}, across the six benchmarks. 
SimpleBaseline employs Resnet-152 as its backbone, while TransPose and TokenPose employ HRNet-W48. It is important to note that the backbone parameters and FLOPs count for these methods have not been included.

The results clearly demonstrate that both our methods outperform all other convolutional and transformer-based methods across all benchmarks, except for MPII. The performance gap is especially pronounced on the more challenging datasets. Notably, even ViTPose-L and ViTPose-H, which are trained with a shared encoder across multiple datasets and have three times and nearly seven times more GFLOPs than our method, respectively, only marginally surpass our approach on MS-COCO and MPII. In contrast, our methods outperformed both variants on all other datasets with significantly lower computational costs.

Our experiments show that \methodvit\ without patch selection performs well, but is computationally expensive. However, our proposed joint-token-based patch selection method (\methodvit/JT) and our previous neighboring (\methodvit/N) and skeleton (\methodvit/S) based patch selection methods from ~\cite{EViTPose} significantly reduce computational costs while maintaining high accuracy. For instance, we achieve a reduction of 30\% to 44\% in GFLOPs with a slight drop in accuracy ranging from 1.1\% to 3.5\% for COCO, 0\% to 0.6\% for MPII, and 0.7\% to 3.5\% for OCHuman.
We can also control the drop in accuracy by changing the number of selected patches. The trade-off between performance and computational complexity for \methodvit/N and \methodvit/JT and a run-time comparison are presented in the Appendix.

For \method, the version \method/C, which uses the classical CNN heat-map decoder is computationally inefficient. However, \method/PS, which uses the proposed pixel shuffle-based efficient decoder, reduces the GFLOPs by more than half without sacrificing performance. 
To further improve efficiency, we can use only the last feature map (i.e. \method/LF) instead of hierarchical feature maps, but this results in a slight drop in performance. Most of the compared methods follow this approach. Similarly, to prioritize efficiency over performance, we can use the smaller variant (\method-S), otherwise, we can use the base variant (\method-B). Besides, we can use the regression decoder to gain 28\% decrease in computation compared to the heat-map decoder although it results in a sub-optimal performance (see Appendix).

\textbf{Ablation:} To assess the impact of the unified skeletal representation, we compared the performance of \method\ trained on a single dataset and the unified \method\ trained on multiple datasets. The results shown in Table~\ref{tab:unified_vs_single} demonstrate that the performance of UniTransPose significantly improves with the use of the unified skeleton representation
and training on multiple datasets. Nonetheless, \method\ trained on a single dataset outperforms current state-of-the-art methods. Furthermore, we evaluate the impact of JAGL attention in contrast to exclusively using local attention (see Table~\ref{tab:jagl_ablation}). The results show that local attention with a cross window outperforms local attention with a shifted window for the most part. However, the incorporation of global attention through the joint tokens in JAGL enhances performance, underscoring the efficacy of propagating global information to patch tokens.

We also conducted a run-time comparison (measured in frames per second, FPS) among \methodvit, ViTPose, and TokenPose. The results in Appendix (Table~\ref{fig:runtime_analysis}) show that our Joint-Token-based Patch Selection method (\methodvit-B/JT) achieves an 88\% reduction in GFLOPs and 10$\times$ 
increase in FPS with respect to ViTPose-H, with a minimal drop in accuracy of up to 2.9\%.
Please refer to the supplementary for further details regarding the implementation, model variants, and more ablation experiments.

\begin{table*}
  \centering
  \caption{Comparison of the proposed methods and other state-of-the-art methods on the MS COCO val set, AI Challenger val set, CrowdPose test set, MPII val set, JRDB-Pose val set, and OCHuman test set. All of the methods follow a top-down approach with heat-map decoding. LiteHRNet~\cite{Yu2021LiteHRNetAL}, a lightweight and less accurate pose estimation network, is used to guide the first two patch selection methods (\methodvit-B/N and \methodvit-B/S). UniTransPose\ employs three variants of heat-map decoders: a classical CNN heat-map decoder (UniTransPose/C), an efficient pixel-shuffle-based decoder (UniTransPose/PS), and a CNN decoder that uses only the last feature map (UniTransPose/LF). On the encoder side, there are two variants: the base (i.e. UniTransPose-B) and the smaller version (i.e. UniTransPose-S).} 
  \label{tab:all}
  \begin{tabular}{@{}l|ccc|ccccc|c@{}}
    \toprule
    Model &  Input Size & Params & FLOPs & \multicolumn{1}{c}{MS-COCO} & \multicolumn{1}{c}{AI Challenger} & \multicolumn{1}{c}{CrowdPose} &  \multicolumn{1}{c}{MPII} & \multicolumn{1}{c}{JRDB-Pose} & \multicolumn{1}{c}{OCHuman} \\
    \cmidrule(lr){5-5} \cmidrule(lr){6-6} \cmidrule(lr){7-7} \cmidrule(lr){8-8} \cmidrule(lr){9-9} \cmidrule(lr){10-10} 
    & & & & mAP & mAP & mAP & PCKh & mAP & mAP \\
    % Model & Backbone & Input Resolution & Params & GFLOPs & AP & AR \\
    \midrule \midrule
8-stage Hourglass & $256 \times 192$ & 25M & 14.3G & 66.9 & - & 65.2 & - & - & -  \\ 
SimpleBaseline & $256 \times 192$ & 69M & 15.7G & 72.0 & 29.9  & 60.8 & 89.0 & - & 58.2  \\
HRNet-W48  & $256 \times 192$ & 64M & 14.6G & 75.1 & 33.5 & - & 90.1 & 42.4 & -   \\
HRNet-W48  & $384 \times 288$ & 64M & 32.9G & 76.3 & - & -  & - & - & 61.6 \\
TransPose-H/A6  & $256 \times 192$ & 18M & 21.8G & 75.8 & - & 71.9 & 92.3 & - & - \\
TokenPose-L/D24  & $256 \times 192$ & 28M & 11.0G & 75.8 & -  & - & - & - & - \\
HRFormer-B  & $256 \times 192$ & 43M & 12.2G & 75.6 & 34.4 & 72.4 & - & - & 49.7 \\
ViTPose-B & $256 \times 192$ & 90M & 18.0G & 77.1 & 32.0 & - & 93.3 & - & 87.3 \\
ViTPose-L  & $256 \times 192$ & 309M & 59.8G & {78.7} & 34.5 & - & {94.0} & - & 90.9 \\
ViTPose-H & $256 \times 192$ & 638M & 122.8G & {79.5} & 35.4 & - & {94.1} & - &  90.9  \\
% \midrule
\midrule
EViTPose-B & $256 \times 192$ & 90M & 19.8G & {77.6} & 36.6 & 76.3 & 92.4 & 73.9 & 93.0 \\
EViTPose-B/N~\cite{EViTPose} & $256 \times 192$ &  90M & 11.1G & 74.1 & - & - & 91.8 & - & 89.5 \\
EViTPose-B/S~\cite{EViTPose} & $256 \times 192$ &  90M & 13.3G & 75.0 & - & - & 92.1 &  -  & 90.1 \\
EViTPose-B/JT & $256 \times 192$ & 90M & 13.7G & 76.5 & 35.0 & 74.5 & 92.5 & 73.9 & 92.3  \\
\midrule
% \midrule
UniTransPose-S/C  & $256 \times 192$ & 58M & 23.5G  & 75.9 &  34.0  & 74.9 & 91.6 & 72.2 & 85.5   \\
UniTransPose-S/PS  & $256 \times 192$ & 34M & 10.0G  & 76.2  & 34.4 & 74.8 & 91.6 & 72.0 & 86.7  \\
UniTransPose-S/LF  & $256 \times 192$ & 32M & ~~5.7G  & 73.4   & 32.0    & 73.6  & 91.4 & 70.9 & 84.9   \\
\midrule
UniTransPose-B/PS  & $256 \times 192$ & 84M & 18.1G  & {78.0} & {37.7}  & {78.2} & {92.5} & {73.7} & {93.5}  \\
UniTransPose-B/LF  & $256 \times 192$ & 80M & 13.6G  & 77.2 & 35.4 & 76.6 & 92.4 & 72.8 & 91.6 \\
% \midrule
UniTransPose-B/PS  & $384 \times 288$ & 84M & 41.0G & {79.3} & {42.1} & {79.3} & {93.1} & {74.7} & {93.9} \\ 
% \midrule
% \midrule

    \bottomrule 
    
  \end{tabular}
  
\end{table*}

\begin{table*}
  \centering
  \caption{Performance comparison of \method\ with and without the unified skeletal representation, trained on single versus multiple datasets. The results indicate significant improvement in performance with the use of the unified training, while the \method\ trained on a single dataset performs comparably or outperforms state-of-the-art methods, particularly on large-scale datasets.}
  \label{tab:unified_vs_single}
  \begin{tabular}{@{}lcccccccccccc@{}}
    \toprule
    Model & Input Size & Unified & \multicolumn{2}{c}{MS-COCO} & \multicolumn{2}{c}{AI Challenger} & \multicolumn{2}{c}{CrowdPose} &  \multicolumn{2}{c}{OCHuman} &  \multicolumn{2}{c}{JRDB-Pose} \\
    \cmidrule(lr){4-5} \cmidrule(lr){6-7} \cmidrule(lr){8-9} \cmidrule(lr){10-11} \cmidrule(lr){12-13} 
    & & Training & AP          & AR         & AP           & AR       & AP           & AR   & AP           & AR  & AP           & AR\\
    % Model & Backbone & Input Resolution & Params & GFLOPs & AP & AR \\
    \midrule
    \method-B/PS & $256\times 192$ & &            76.1 & 81.4 & 35.6 & 38.8 & 69.9 & 79.1 & 60.7 & 65.1 & 65.9 & 70.4 \\ \midrule
    \method-B/PS & $256\times192$ & \checkmark & 78.0 & 83.2 & 37.7 & 41.6 & 78.2 & 86.6 & 93.5 & 94.7 & 73.7 & 77.0\\
  \bottomrule
  \end{tabular}
  
\end{table*}

\begin{table*}[h!]
  \centering
  \caption{Performance comparison of \method\ with local attention variants and JAGL attention. The results indicate significant improvement in performance with the use of the JAGL attention.}
  \label{tab:jagl_ablation}
  \begin{tabular}{@{}llccccccccc@{}}
    \toprule
    Model & Attention Type & \multicolumn{2}{c}{MS-COCO} & \multicolumn{2}{c}{AI Challenger} & \multicolumn{2}{c}{CrowdPose} &  \multicolumn{2}{c}{OCHuman} &  \multicolumn{1}{c}{MPII} \\
    \cmidrule(lr){3-4} \cmidrule(lr){5-6} \cmidrule(lr){7-8} \cmidrule(lr){9-10} \cmidrule(lr){11-11} 
    &  & AP          & AR         & AP           & AR       & AP           & AR   & AP           & AR  & PCKh \\
    % Model & Backbone & Input Resolution & Params & GFLOPs & AP & AR \\
    \midrule
    \method-B/PS & Local (Shifted Window) \cite{Liu2021SwinTH} &            76.7 & 82.0 & 35.2 & 39.4 & 76.0 & 84.3 & 89.6 & 91.2 & 92.3  \\ \midrule
    \method-B/PS & Local (Cross Window) \cite{Dong2022CSWinTA} &              77.1 & 82.1 & 36.6 & 40.3 & 76.9 & 84.8 & 88.3 & 90.1 & 92.1  \\ \midrule
    \method-B/PS &  JAGL (Local + Global) & 78.0 & 83.2 & 37.7 & 41.6 & 78.2 & 86.6 & 93.5 & 94.7 & 92.5 \\
  \bottomrule
  \end{tabular}
  
\end{table*}

\begin{figure}[tb]
    \includegraphics[width=0.99\columnwidth,clip=true, trim = 0 30 0 70]{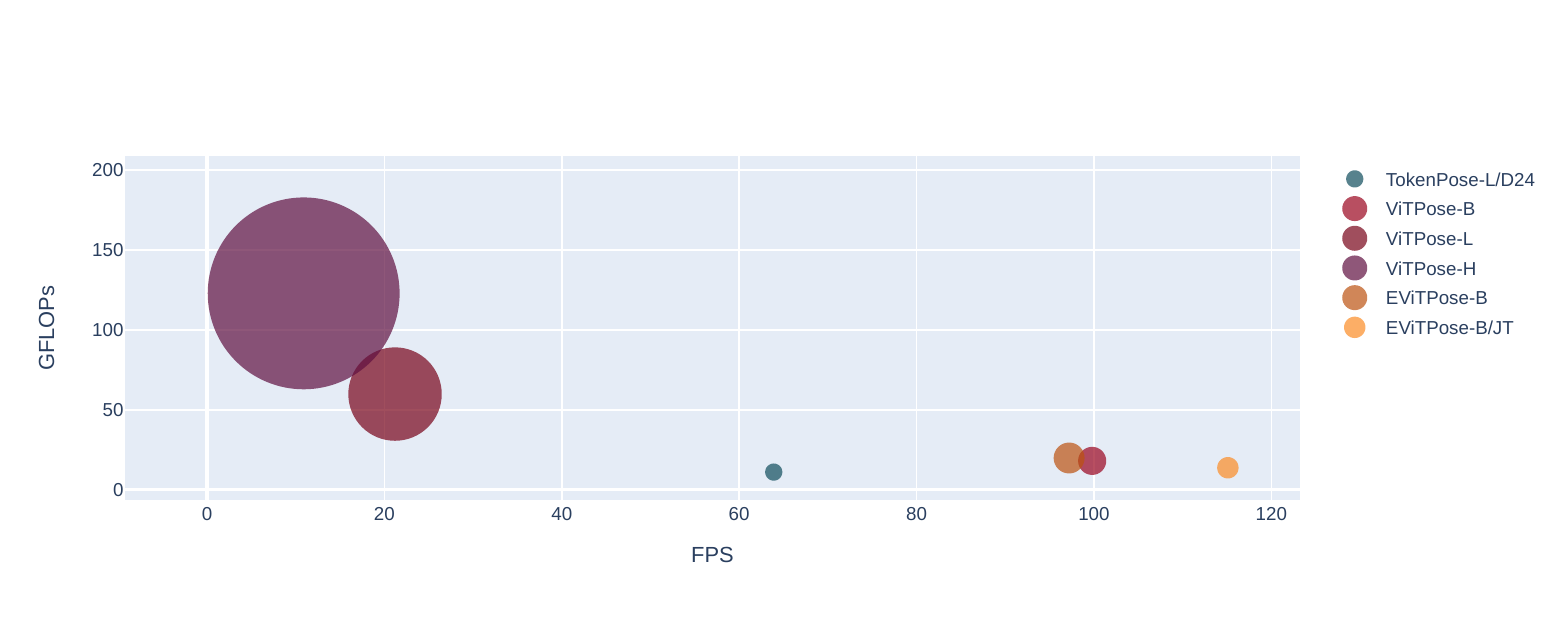}
  \caption{\textbf{Runtime (FPS) vs GFLOPs comparison --}
  The Joint-Token-based Patch Selection method (\method-B/JT) achieves an 88\% reduction in GFLOPs and a 10$\times$ (955\%) increase in FPS compared to ViTPose-H, with a minimal accuracy drop of up to 2.9\%.
  % The x-axis shows the FPS, and the y-axis shows the GFLOPs.
  }
  \label{fig:runtime_analysis}
 
\end{figure}

\section{Conclusion}
This work presented two transformer-based approaches to 2D human pose estimation addressing challenges with state-of-the-art methods. The first method, EViTPose, is a Vision Transformer-based network that employs patch selection methods to substantially reduce computational complexity without compromising accuracy. The proposed patch selection methods leverage fast pose estimation networks and learnable joint tokens to achieve a remarkable reduction in GFLOPs (30\% to 44\%) across six benchmark datasets, with only a marginal decline in accuracy (0\% to 3.5\%). The second approach, UniTransPose, introduces a multi-scale transformer with local-global attention coupled with an efficient sub-pixel CNN decoder and a simple key-point regressor. Both methods unify joint annotations from multiple datasets, improving generalization across different benchmarks and outperforming previous state-of-the-art methods in terms of both accuracy and computational complexity.

\section*{Acknowledgments}
% This should be a simple paragraph before the References to thank those individuals and institutions who have supported your work on this article.
The authors thank Carolina Pacheco, Yutao Tang, Darshan Thaker, and Yuyan Ge for their valuable input and feedback throughout the development of this work. This research is based upon work supported in part by NIH grant 5R01NS135613, NSF grant 2124277 and IARPA grant 2022-21102100005. The views and conclusions contained herein are those of the authors and should not be interpreted as necessarily representing the official policies, either expressed or implied, of NIH, NSF, IARPA, or the U.S. Government. The U.S. Government is authorized to reproduce and distribute reprints for governmental purposes notwithstanding any copyright annotation therein.

%%%%%%%%% REFERENCES

% \begin{thebibliography}
% \bibliographystyle{IEEEtran}

% \end{thebibliography}
\bibliography{bib.bib}{}
\bibliographystyle{IEEEtran}

% \newpage

\section{Biography Section}
\vskip 0pt plus -1fil
 

% \vspace{-60pt}
\begin{IEEEbiography}[{\includegraphics[width=1in,height=1.25in,clip,keepaspectratio]{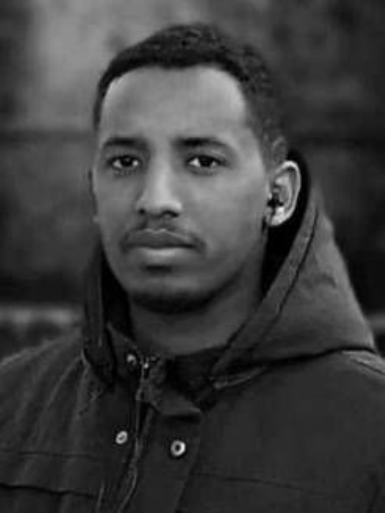}}]{Kaleab A. Kinfu} (Student Member, IEEE) received the BS degree in Computer Science
from Addis Ababa University and the MS degrees in Computer Science Engineering, Image Processing and Computer Vision, Computer Science, and Biomedical Engineering, from Pazmany Peter Catholic University, Autonomous University of Madrid, University of Bordeaux, and the Johns Hopkins University, respectively. He is currently working toward the PhD degree with
the Computer and Information Science Department, University of Pennsylvania. His research interests include
human pose estimation, motion analysis and synthesis, robustness and generalization in machine learning.
\end{IEEEbiography}
\vskip 0pt plus -1fil
\begin{IEEEbiography}[{\includegraphics[width=1in,height=1.25in,clip,keepaspectratio]{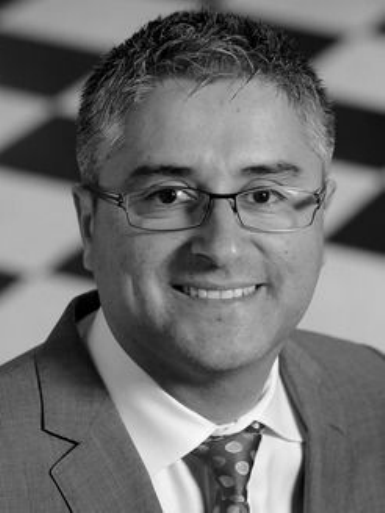}}]{René Vidal}
(Fellow, IEEE) received the BS degree in electrical engineering (valedictorian)
from the Pontificia Universidad Catolica de Chile, in 1997, and the MS and PhD degrees in electrical engineering and computer science from the University of California at Berkeley, in 2000 and 2003, respectively. From 2004-2022 he was a Professor of Biomedical Engineering and the Director of the Mathematical Institute for Data Science (MINDS) at The Johns Hopkins University. He is currently the Rachleff University Professor of Electrical and Systems Engineering and Radiology and the director of the Center for Innovation in Data Engineering and Science (IDEAS) at the University of Pennsylvania. He is co-author of the book “Generalized Principal Component Analysis” (Springer 2016), co-editor of the book “Dynamical Vision” (Springer 2006) and co-author of more than 400 articles in machine learning, computer vision, signal and image processing, biomedical image analysis, hybrid systems, robotics and control. He has been associate editor in chief of the IEEE Transactions on Pattern Analysis and Machine Intelligence and Computer Vision and Image Understanding, associate editor or guest editor of Medical Image Analysis, the IEEE Transactions on Pattern Analysis and Machine Intelligence, the SIAM Journal on Imaging Sciences, Computer Vision and Image Understanding, the Journal of Mathematical Imaging and Vision, the International Journal on Computer Vision and Signal Processing Magazine. He has
received numerous awards for his work, including the 2021 Edward J. McCluskey Technical Achievement Award, the 2016 D’Alembert Faculty Fellowship, the 2012 IAPR J.K. Aggarwal Prize, the 2009 ONR Young Investigator Award, the 2009 Sloan Research Fellowship and the 2005 NSF CAREER Award. He is a fellow of the ACM, AIMBE, IAPR and IEEE, and a member of the SIAM.
\end{IEEEbiography}

\newpage

{\appendices
\section{Implementation Details}
To achieve a balance between efficiency and accuracy, we have developed two versions of our model, namely \method-S (Small) and \method-B (Base), as outlined in Table~\ref{tab:variants}. These variants were created by changing the base channel dimension $C$ and the JAGL block number in each stage. It's worth mentioning that each JAGL block comprises two layers of local attention followed by global attention. 

In the smaller variant, the four stages have 1,10,1, and 1 JAGL blocks respectively, with a base channel dimension of 64. In the base variant, the four stages have 1,2,12, and 1 JAGL blocks respectively, with a channel dimension of 96. Both variants maintain an expansion ratio of 4 for each MLP and a stripe width of 1,2,7, and 7 for the four stages of local attention, respectively. Furthermore, the small variant has 2,4,8, and 16 heads for the four stages in local attention, while the base variant has 4,8,16, and 32 heads, respectively. In global attention, the small variant has 1,2,4, and 8 heads for the four stages, while the base variant has 2,4,8, and 16 heads.

Additionally, both variants utilize a two-layer MLP with varying channel dimensions as a key-point regressor. The small variant has an input channel dimension of 512 and layer channel dimensions of 128 and 2. The base variant has an input channel dimension of 768 and layer channel dimensions of 192 and 2.

\begin{table*}[bhp]
  \centering
  \caption{Comparison of architecture details between the smaller (\method-S) and base (\method-B) variants, covering the base channel dimension, number of JAGL blocks, stripe width in local attention, head numbers in local and global attention for each of the four stages, as well as the channel dimensions used in key-point regression. }
  \label{tab:variants}
  \begin{tabular}{@{}lccccccc@{}}
    \toprule
    Model &  Channel & JAGL Blocks & \multicolumn{2}{c}{Local Attention} & \multicolumn{1}{c}{Global Attention}  & \multicolumn{2}{c}{key-point regressor}\\
    \cmidrule(lr){4-5}  \cmidrule(lr){7-8}
      & Dim.    &   & Stripes Width & \#Heads &  \#Heads & Input Chan. & Layer Chan. \\
    \midrule
    \method-S & 64 & $[1,1,10,1]$ & $[1,2,7,7]$ & $[2,4,8,16]$  & $[1,2,4,8]$ & 512 & $[128, 2]$ \\
    \method-B & 96 & $[1,2,12,1]$ & $[1,2,7,7]$ & $[4,8,16,32]$  & $[2,4,8,16]$ & 768 & $[192, 2]$ \\ 
  \bottomrule
  \end{tabular}
  
\end{table*}

\section{Detailed Experimental Settings}
\label{sup:experiment_details}
The training methodology largely follows mmpose\footnote{https://github.com/open-mmlab/mmpose, Apache License 2.0}, wherein a default input image resolution of $256\times 192$ is adopted for both model variants. In order to optimize GPU usage, when training at a resolution of $384\times 288$ or changing the decoder variants, the model trained at $256\times 192$ with the pixel-shuffle-based decoder is fine-tuned rather than being trained from scratch.

During training with a $256\times 192$ input, an AdamW~\cite{Loshchilov2019DecoupledWD} optimizer is employed for 210 epochs with a learning rate decay by a factor of 10 at the 170-th and 200-th epoch. The batch size is set to $48$, with an initial learning rate of $5e-6$, weight decay of $0.01$, and gradient clipping with a maximum norm of $1$. Most of the augmentation and regularization strategies of mmpose are incorporated into training, and $\lambda$ — which controls how much weight is given to key-point regression loss — is set to $10e-2$.

For fine-tuning, an AdamW~\cite{Loshchilov2019DecoupledWD} optimizer is used for 30 epochs with a constant learning rate of $9e-7$, weight decay of $10e-6$, and the same data augmentation and regularizations as before.

\section{Ablation Experiments}
\label{sup:ablation_experiments}

\myparagraph{Trade-off between accuracy and efficiency} In \methodvit, we can control the drop in accuracy by changing the number of patches to be selected. The trade-off between performance and computational complexity for the neighboring~\cite{EViTPose} and joint-token-based patch selection methods is depicted in Figure~\ref{fig:tradeoff}. The neighboring and skeleton patch selection remove irrelevant patches before they are processed by ViT, while the joint-token-based selection method learns to remove them on the fly. Thus, the first two approaches prioritize efficiency over accuracy by removing patches early on. For example, they are effective for addressing the low end range in Figure~\ref{fig:tradeoff}, where the joint-token-based selection method performs poorly. 

\begin{figure}[tb]
    \includegraphics[width=0.99\columnwidth,clip=true, trim = 0 29 0 12]{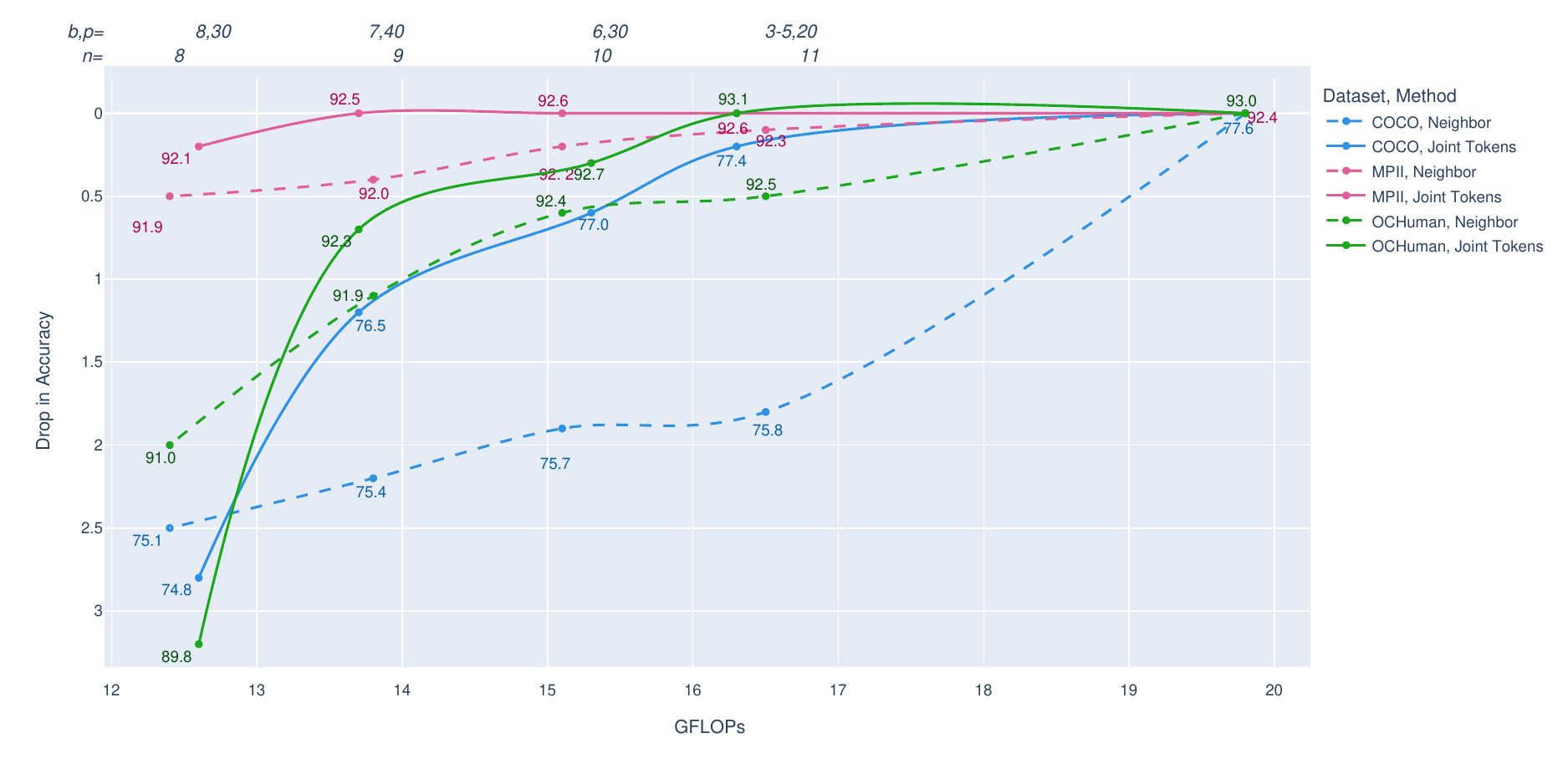}
  \caption{\textbf{Trade-off between accuracy and GFLOPs on three benchmarks: COCO, MPII, and OCHuman --} 
  % The x-axis shows the GFLOPs, and the y-axis shows the drop in accuracy. 
  The performance of \method-B with two patch selection methods: Neighbors (dashed line) and Joint-Token-based (solid line).
  $n$ denotes to the number of neighbors selected and $b,p$ refers to $p$ number of patches that are removed at block $b$ in the Joint Tokens method.}
  \label{fig:tradeoff}
 
\end{figure}

\myparagraph{key-point regression versus Heat-Map Decoding} In both methods, we adopt a flexible approach in our network training by utilizing both decoding options -- key-point regression and heat-map decoding -- to strike a balance between accuracy and efficiency. This flexibility allows users to select either the key-point regressor for quick estimates or the heat-map decoder for more accurate predictions during inference. In the main paper, we presented the results using the heat-map decoding approach. Here, we present a comparison of \method's simple key-point regression decoding 
% with other methods that employ similar decoding strategies, along 
with the efficient heat-map decoding approach, as summarized in Table~\ref{tab:keypoint_regression}. The key-point regression simply uses the direct regression with Smooth $L_1$ loss. However, the utilization of advanced techniques such as Residual Log-likelihood Estimation~\cite{Li2021HumanPR} for regression could potentially enhance the accuracy of the key-point regression model.

\begin{table*}
  \centering
  \caption{Comparison of key-point decoding options - key-point regression versus heat-map decoding - for a balance between accuracy and efficiency. key-point regression (\method/KR) provides quick estimates but is less robust, while heat-map decoding (\method/PS) offers more accurate predictions. Note that the heat-map decoder used here is the efficient pixel-shuffle-based decoder.}
  \label{tab:keypoint_regression}
  \begin{tabular}{@{}lccccccccccccc@{}}
    \toprule
    Model & Input Size & Params & FLOPs & \multicolumn{2}{c}{MS-COCO} & \multicolumn{1}{c}{MPII} & \multicolumn{2}{c}{CrowdPose} &  \multicolumn{2}{c}{OCHuman} &  \multicolumn{2}{c}{JRDB-Pose} \\
    \cmidrule(lr){5-6} \cmidrule(lr){7-7} \cmidrule(lr){8-9} \cmidrule(lr){10-11} \cmidrule(lr){12-13} 
    & & & &  AP          & AR         & PCKh       & AP           & AR   & AP           & AR  & AP           & AR\\
    % Model & Backbone & Input Resolution & Params & GFLOPs & AP & AR \\
    % PointSetNet & - & - & - & 65.7 & - & \\
    \midrule
    PRTR~\cite{Li2021PoseRW} & $384 \times 288$ & 42M & 11.0G & 68.2 & 76.0 & 88.2 & - & - & - & - & - & - \\
    \method/KR & $256\times192$ & 78M & 13.0G & 68.0 & 77.2 & 91.2 & 60.6 & 82.4 & 83.1 & 89.5 & 64.4 & 73.3 \\     \midrule

    \method/PS & $256 \times 192$ & 84M & 18.1G & 78.0 & 83.2 & 92.5 & 78.2 & 86.6 & 93.5 & 94.7 & 73.7 & 77.0 \\
  \bottomrule
  \end{tabular}
\end{table*}

\section{Visual Results}
\label{sup:visual_results}
The pose estimation results of \method\ on a few randomly chosen samples from the MS-COCO dataset are depicted in Figure \ref{fig:coco_results} to demonstrate its effectiveness. The accuracy of the results is apparent from the illustrations, which exhibit challenging scenarios like heavy occlusion, varying postures, and scales.

\begin{figure*}
    \centering
    \includegraphics[width=\textwidth]{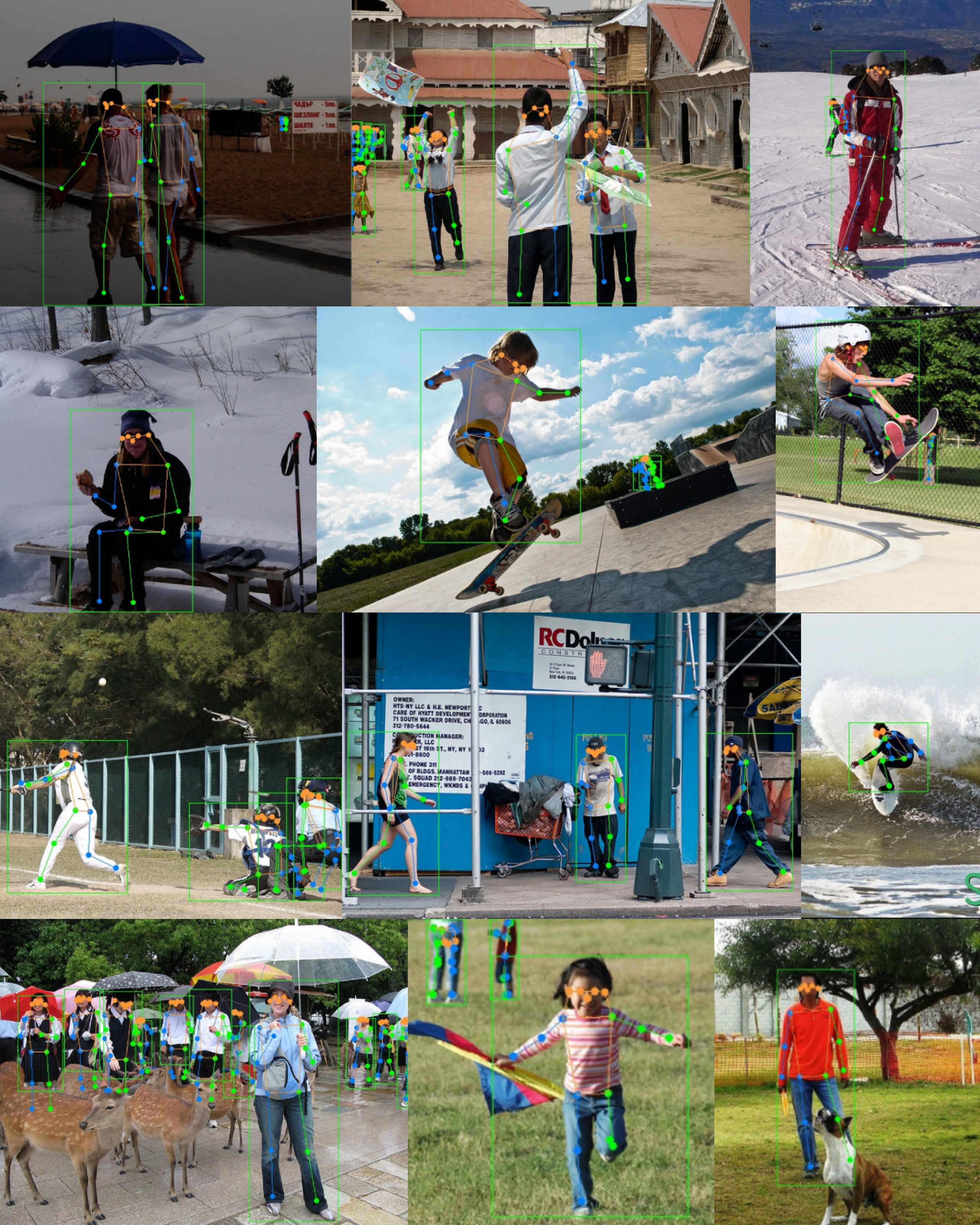}
    \caption{Visualization of pose estimation results obtained by \method-B/PS on a few MS-COCO val images.}
    \label{fig:coco_results}
\end{figure*}

}

\end{document}